\documentclass{article}

\usepackage[final, dandb]{neurips_2025}

\usepackage[T1]{fontenc}    
\usepackage[hidelinks]{hyperref}       
\usepackage{url}            
\usepackage{booktabs}       
\usepackage{amsfonts}       
\usepackage{nicefrac}       
\usepackage{microtype}      
\usepackage[dvipsnames]{xcolor}

\usepackage{tabularx}

\usepackage{array}
\newcolumntype{C}[1]{>{\centering\arraybackslash}p{#1}}  

\usepackage{graphicx}  
\usepackage{caption}   
\usepackage{subcaption}
\usepackage{adjustbox}

\usepackage{booktabs}  
\usepackage{array}     
\usepackage{amssymb} 
\usepackage{pifont}  
\usepackage{float}
\usepackage{multirow}
\usepackage{makecell}
\usepackage{colortbl}
\usepackage{diagbox}
\usepackage{amsmath}

\usepackage{float} 

\usepackage{xfrac}

\usepackage{marvosym} 
\usepackage{footmisc}

\usepackage{listings}
\usepackage[most]{tcolorbox}
\usepackage{verbatim}
\tcbuselibrary{listingsutf8}

\DeclareMathOperator*{\argmax}{\arg\!\max}

\title{A2Seek: Towards Reasoning-Centric Benchmark for Aerial Anomaly Understanding}

\author{%
  Mengjingcheng Mo$^{1,2}$,
  Xinyang Tong$^{1}$,
  Mingpi Tan$^{1}$, 
  Jiaxu Leng$^{1,2}$\thanks{Corresponding authors},\\
  \textbf{Jiankang Zheng}$^{1,2}$,
  \textbf{Yiran Liu}$^{1}$,
  \textbf{Haosheng Chen}$^{1}$,
  \textbf{Ji Gan}$^{1,2}$,
  \textbf{Weisheng Li}$^{1}$,
  \textbf{Xinbo Gao}$^{1,2}$\footnotemark[\value{footnote}]\\
  $^1$School of Computer Science and Technology, \\
  Chongqing University of Posts and Telecommunications, Chongqing, China \\
  $^2$Chongqing Institute for Brain and Intelligence, Guangyang Bay Laboratory, Chongqing, China \\
  \texttt{\{lengjx,gaoxb\}@cqupt.edu.cn}
}

\begin{document}

\maketitle

\begin{abstract}

While unmanned aerial vehicles (UAVs) offer wide-area, high-altitude coverage for anomaly detection, they face challenges such as dynamic viewpoints, scale variations, and complex scenes. Existing datasets and methods, mainly designed for fixed ground-level views, struggle to adapt to these conditions, leading to significant performance drops in drone-view scenarios.
To bridge this gap, we introduce A2Seek (Aerial Anomaly Seek), a large-scale, reasoning-centric benchmark dataset for aerial anomaly understanding. This dataset covers various scenarios and environmental conditions, providing high-resolution real-world aerial videos with detailed annotations, including anomaly categories, frame-level timestamps, region-level bounding boxes, and natural language explanations for causal reasoning. 
Building on this dataset, we propose A2Seek-R1, a novel reasoning framework that generalizes R1-style strategies to aerial anomaly understanding, enabling a deeper understanding of ``Where'' anomalies occur and ``Why'' they happen in aerial frames.
To this end, A2Seek-R1 first employs a graph-of-thought (GoT)-guided supervised fine-tuning approach to activate the model's latent reasoning capabilities on A2Seek. Then, we introduce Aerial Group Relative Policy Optimization (A-GRPO) to design rule-based reward functions tailored to aerial scenarios. Furthermore, we propose a novel ``seeking'' mechanism that simulates UAV flight behavior by directing the model's attention to informative regions.
Extensive experiments demonstrate that A2Seek-R1 achieves up to a 22.04\% improvement in AP for prediction accuracy and a 13.9\% gain in mIoU for anomaly localization, exhibiting strong generalization across complex environments and out-of-distribution scenarios. Our dataset and code are released at \url{https://2-mo.github.io/A2Seek/}.
\end{abstract}

\section{Introduction}
\label{introduction}
\begin{figure*}[t!]
  \centering
    \includegraphics[width=\linewidth]{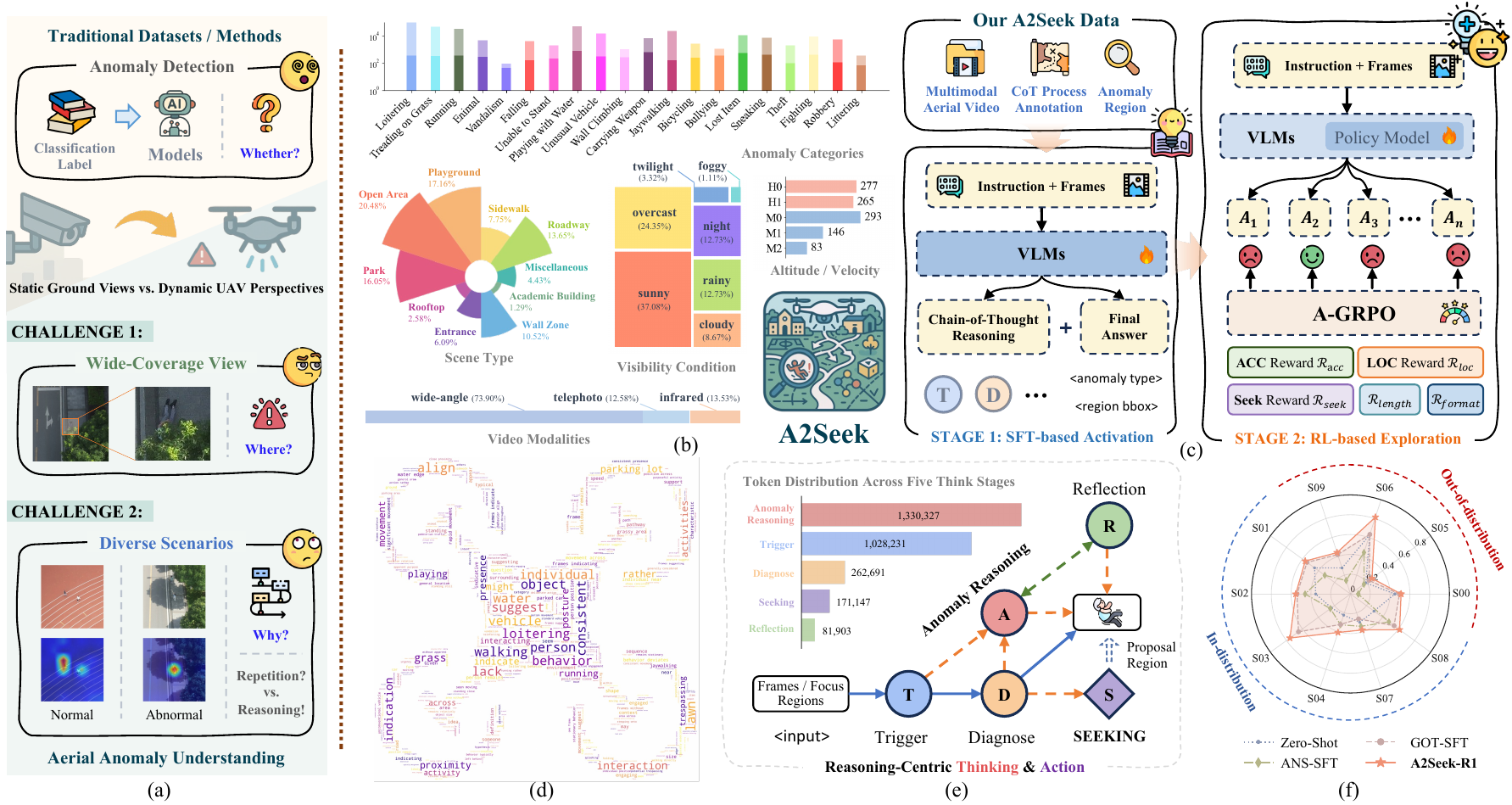}
   \caption{Overview of the A2Seek Benchmark. \textbf{(a)} Challenges in aerial anomaly detection. Traditional methods rely on static surveillance views and focus mainly on classification, making it difficult to answer ``Where'' and ``Why'' anomalies occur under dynamic UAV perspectives. \textbf{(b)} Dataset statistics on multiple dimensions. \textbf{(c)} Reasoning pipeline. The method consists of two stages: SFT (supervised fine-tuning) for reasoning activation, and RL (reinforcement learning) for dynamic reasoning. \textbf{(d)} High-frequency word of dataset. \textbf{(e)} Reasoning process. The framework integrates multiple reasoning stages (Trigger, Diagnose, Reasoning, Reflection and Seeking), emphasizing reasoning-driven anomaly understanding. \textbf{(f)} Performance comparison.}
   \vspace{-1em}
   \label{fig:intro}
\end{figure*}

Traditional anomaly detection \cite{lu2013abnormal, liu2018ano_pred, cao2023new} relies on fixed-view cameras and primarily focuses on anomaly classification, offering limited semantic interpretation. Their static perspectives and narrow fields of view significantly limit their effectiveness in monitoring large and dynamic environments~\cite{zhu2024advancing}.
With the rapid advancement of unmanned aerial vehicle (UAV) technology, aerial surveillance has emerged as a powerful paradigm for wide-area anomaly detection. 
Drone‐view footage introduces frequent viewpoint shifts, scale changes, complex backgrounds and occlusions, as well as environmental disturbances (lighting, weather, moving shadows) \cite{cao2021visdrone, du2018unmanned}. 
Crucially, anomalous regions in aerial scenes are often subtle, spatially sparse, and occupy only a small portion of the field of view, making them difficult to perceive. 
Even when alarms are triggered, human observers often struggle to localize these subtle, spatially sparse events. 
Moreover, traditional models rely on stable spatial priors invalidated when a drone's pose constantly changes. 
Consequently, ground‐view approaches fail to generalize to aerial data, exhibiting poor robustness to dynamic scenes and variable spatial distributions.
As Figure~\ref{fig:intro} illustrates, anomaly detection in drone-view footage requires precise spatial localization (``Where is the anomaly?'') and semantic interpretation (``Why is it an anomaly?'').
Drone-captured videos often span large, dynamic environments, where subtle anomalies can be easily overlooked or misinterpreted without robust contextual reasoning. 
These factors demand models capable of adaptively focusing on critical regions and abstracting high-level reasoning to explain anomalies. 
In the following, we distill these into two core problems critical for robust aerial anomaly understanding.

\textbf{CHALLENGE 1: Spatial Localization in Wide-Area, Dynamic Scenes. }  
Traditional ground-based anomaly detection typically relies on fixed-view cameras, where anomalies are often prominent and relatively easy to identify. However, aerial anomaly detection is fundamentally different. Drone-view videos cover expansive and dynamic scenes with complex backgrounds and frequent motion. Anomalies are often small, sparse, and easily obscured by cluttered surroundings, making them difficult to detect even when anomalies are flagged (\textit{e.g.}, a person falling in Figure \ref{fig:intro}a). Therefore, aerial anomaly detection requires not only determining whether an event is anomalous but also providing accurate localization of the anomalous objects to support anomaly judgment.

\textbf{CHALLENGE 2: Semantic Generalization Across Diverse Aerial Contexts. }  
Aerial anomaly detection faces significant scene dependency and distributional shifts, as the same behavior (\textit{e.g.} running) may exhibit vastly different visual characteristics and interactions across environments (\textit{e.g.} playgrounds or academic buildings). Relying solely on feature classification or reconstruction errors fails to capture this diversity and hinders reliable generalization to unseen scenarios. To address this, models must deeply understand scene context and behavioral semantics to effectively tackle the generalization challenge. 


Recent advances in multimodal large language models (MLLMs) \cite{achiam2023gpt, chen2024internvl, bai2025qwen2, li2024llavanext-strong} have demonstrated significant potential in cross-modal alignment and semantic understanding, particularly in complex reasoning scenarios \cite{lv2024video, wang2024internvideo2, zhang2024holmes}. However, their application to aerial anomaly understanding remains underexplored due to two major limitations: (1) the lack of large-scale multimodal aerial datasets with temporal annotations, precise spatial localization, and semantic reasoning explanations, which hinders effective training and evaluation; and (2) the absence of structured reasoning frameworks and adaptive strategies, making it difficult to address diverse anomalies in complex aerial perspectives.
To address the above challenges, we present A2Seek, a reasoning-centric benchmark specifically designed for aerial anomaly understanding. Collected across 10 campus scenes over one year, the dataset spans 23 hours of UAV footage with diverse flight altitudes, speeds, and trajectories, including 3.79 hours of complex anomalies and the rest normal behaviors. It features 542 untrimmed 4K drone videos and over 32k curated keyframes, annotated with fine-grained anomaly labels, spatiotemporal bounding boxes, and structured reasoning graphs. These annotations enable comprehensive evaluation of detection accuracy and reasoning interpretability. To tackle practical challenges such as occlusion and low-light conditions, A2Seek incorporates telephoto footage for high-altitude scenes and infrared modalities for nighttime scenarios, facilitating the detection of subtle or visually ambiguous anomalies.

Building on this benchmark, we propose A2Seek-R1, a novel reinforcement fine-tuning framework designed to enhance the reasoning capabilities of models for aerial anomaly understanding. A2Seek-R1 first employs a graph-of-thought (GoT)-guided supervised fine-tuning (SFT) approach, which activates the model's latent reasoning capabilities by leveraging structured reasoning annotations in the A2Seek dataset. 
These annotations consist of optional stages, including trigger, diagnosis, reasoning, reflection, and seeking, effectively guiding the model to handle anomalies of varying complexity in a progressive manner.  
Among them, seeking is set as a potential region of interest for the model in video frames with insufficient information, such as blurry or occluded images, thus achieving a new type of seeking mechanism that simulates the flight behavior of unmanned aerial vehicles, enabling the model to dynamically focus on specific regions of interest.
Second, it introduces a tailored extension of Group Relative Policy Optimization (GRPO), termed A-GRPO, specifically designed for aerial anomaly understanding. A-GRPO extends the original accuracy and format function rewards by incorporating localization and seeking rewards. Localization rewards enhance the model's spatial understanding of anomaly regions, while seeking rewards focus on aligning the model's predictions with human annotations of anomaly candidate areas, ensuring the extraction of valuable spatial information for better understanding. 
Additionally, to address the diverse perspectives of drones, a length reward function is introduced to encourage concise responses in simple scenarios and allocate more computational effort to complex situations. By combining these components, A2Seek-R1 achieves precise spatial localization and robust reasoning for aerial anomalies, setting a new benchmark for anomaly understanding in complex, real-world environments.

\textbf{Contributions}: 
\textbf{(1)} We present A2Seek, a large-scale, reasoning-centric benchmark specifically designed for multi-scenario anomaly understanding from aerial perspectives.
\textbf{(2)} We propose A2Seek-R1, a novel multi-stage reinforcement fine-tuning framework that significantly enhances the aerial anomaly understanding capabilities of multimodal foundation models.
\textbf{(3)} This work is the first to simulate UAV motion characteristics in the context of anomaly understanding, enabling models to actively acquire detailed regional information in challenging scenarios. 
\textbf{(4)} Extensive experiments across multiple scenarios validate the superiority of A2Seek-R1. Compared to models trained solely with GoT-SFT, A2Seek-R1 achieves an improvement of 6.72\% in prediction accuracy. 

\section{Related Work}
\label{sec:related}
\textbf{Video Anomaly Detection.} 
Early efforts focused on single-scene datasets \cite{li2013anomaly, lu2013abnormal} using fixed-view RGB cameras for pedestrian anomaly detection. Later datasets \cite{liu2018ano_pred, ramachandra2020street, degardin2021iterative, lv2021localizing} that towards real-world introduced more complex scenes with crowded traffic, yet remained unimodal, fixed-view, and emphasized foreground representation. Larger-scale datasets \cite{sultani2018real, wu2020not, wan2021anomaly, NEURIPS2024_a3c5af1f} improved diversity and duration, but still relied on ground-view perspectives and coarse anomaly labels, lacking spatial localization or causal reasoning.
Methodologically, the field evolved from handcrafted features to learned representations \cite{adam2008robust, zhao2011online, li2013anomaly} and behavior modeling \cite{mehran2009abnormal, li2018resound, kim2009observe}. Recent approaches span reconstruction/prediction-based \cite{hasan2016learning, lv2023unbiased, gong2019memorizing, xu2017detecting, yang2023video, liu2018future}, object-centric \cite{liu2021hybrid, li2021variational, lv2021learning, tian2022pixel, gao2023atta, zhao2024segment}, distribution-aware \cite{hirschorn2023normalizing, tu2024self, micorek2024mulde, leng2024beyond}, and llm-driven paradigms \cite{du2024uncovering, zanella2024harnessing, lv2024video, zhang2024holmes}, with growing emphasis on generalization and scene dependency.
Recent methods \cite{zhu2025vaur1, huang2025vadr1} incorporate chain-of-reasoning mechanisms to enhance anomaly understanding, enabling more interpretable and goal-directed decision-making.
However, existing benchmarks and methods largely overlook aerial-specific challenges such as extreme viewpoint shifts and scale variation. Moreover, the absence of multimodal and reasoning-oriented annotations limits fine-grained analysis. 

\textbf{Aerial Anomaly Understanding.}
Anomaly detection in aerial videos remains underexplored. Early pioneer datasets \cite{bonetto2015privacy, bozcan2021context, jin2022anomaly, tran2023uit} adopt aerial perspectives but provide only frame-level or coarse labels, limiting fine-grained analysis. Existing methods focus on motion cues, including optical flow~\cite{bozcan2021context}, reconstruction-based schemes\cite{le2025hstforu}, or spatiotemporal modeling with 3D CNNs~\cite{tran2023uit} and Transformer~\cite{jin2022anomaly}, but rarely support precise region-level reasoning.
While multimodal large language models (MLLMs) \cite{achiam2023gpt, hu2024minicpm, chen2024internvl, chen2024far, li2024llavanext-strong, yang2024qwen2} have advanced semantic understanding in ground-view tasks, their application to aerial scenarios is limited. Current approaches \cite{bharadwaj2024vanebench, tang2024hawk, zhang2024holmes} often lack explicit, grounded reasoning and rely on post hoc explanations. 
To bridge this gap, we introduce a reasoning-centric aerial anomaly dataset with fine-grained spatial-temporal annotations and dynamic reasoning trajectories. 

\section{The A2Seek Dataset} 
\label{dataset}

\begin{figure*}
  \centering
  \includegraphics[width=\linewidth]{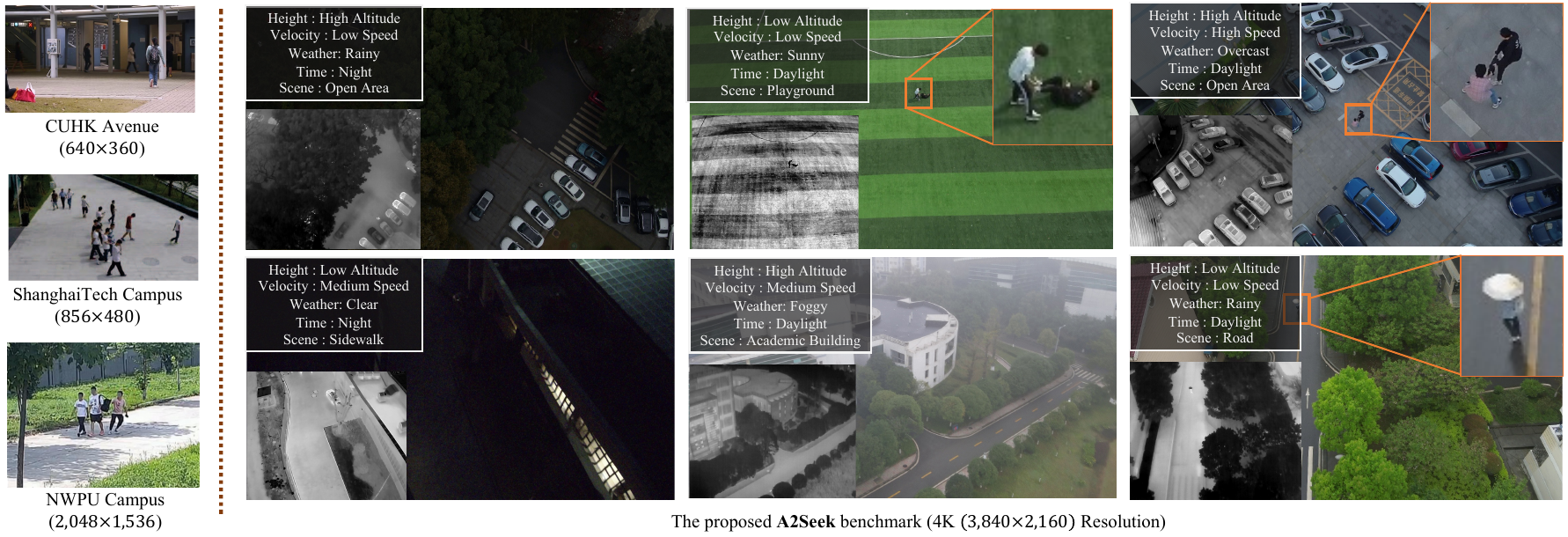}
  \caption{Comparison of scene diversity and complexity. Left: fixed-view surveillance datasets. Right: diverse aerial views in A2Seek.}
  \label{fig:scene_diversity}
  \vspace{-1em}
\end{figure*}

Existing video anomaly detection methods primarily rely on fixed-view ground-based cameras, which are designed for limited fields of view and relatively static backgrounds. These methods face significant limitations when applied to drone-view videos, which involve frequent viewpoint changes, scale variations, dynamic occlusions, and complex environmental disturbances (\textit{e.g.}, lighting changes, weather variations). These factors significantly increase the challenges of spatial localization and semantic generalization of anomaly detection.
To address these challenges, we introduce A2Seek, a reasoning-centric aerial anomaly understanding benchmark. It supports precise spatial localization of anomalies (``Where is the anomaly?'') and in-depth semantic reasoning explanations (``Why is it anomalous?''). The dataset spans diverse real-world scenarios and anomaly types, providing high-resolution RGB and infrared video data with detailed frame-level labels, region-level annotations, and structured natural language reasoning explanations. Figure~\ref{fig:scene_diversity} illustrates the dataset's diversity, showcasing various scenes, altitudes, speeds, weather conditions, and times of day. Unlike traditional ground-based datasets, which focus on static viewpoints and limited environmental variations, A2Seek leverages the dynamic and expansive nature of UAV perspectives, making it a more challenging and realistic benchmark. This benchmark aims to advance research on generalization, robustness, and interpretability in aerial anomaly understanding.

\subsection{Data Collection and Annotation}

The A2Seek dataset was collected using a DJI M30T drone equipped with wide-angle, telephoto, and infrared cameras. Flights were conducted at varying altitudes (10 to 60 meters) and speeds (0 to 20 m/s) to capture diverse aerial perspectives. Trajectory patterns included hovering, linear cruising, curved circling, and area scanning, enabling dynamic viewpoint shifts across scenes. In total, A2Seek comprises 542 untrimmed 4K videos (over 23 hours), recorded across 10 campus environments, covering more than 20 types of anomalous events (\textit{e.g.}, falling, fighting, jaywalking) under varying conditions such as day/night, clear/foggy weather, and so on. \par
The anomaly categories in A2Seek were carefully curated based on the principle of ``potential disruption to campus public safety or order,'' rather than broadly labeling daily activities as abnormal. For instance, running on a playground is not considered anomalous, whereas running in an academic building area is. Similarly, riding a bicycle on the road is normal, but riding on sidewalks is considered anomalous. Drawing from classic VAD datasets \cite{li2013anomaly, lu2013abnormal, luo2017revisit, cao2023new}, we focus on behavior categories with strong relevance to real-world campus safety. 
Anomalies are categorized into three risk levels. High-risk events such as carrying weapons, fighting, and robbery require immediate intervention. Medium-risk events such as running, falling, animal intrusions, and illegal cycling may escalate but do not cause immediate harm. Low-risk events such as loitering, littering, and walking on grass pertain to order maintenance or environmental management.
\par
To ensure high-quality and comprehensive labeling, A2Seek employs a rigorous multi-level annotation framework. This process includes spatiotemporal bounding boxes, fine-grained categories, and structured
reasoning paths. A multi-stage annotation process was designed, encompassing trigger identification, anomaly diagnosis, causal reasoning, reflection, and seeking phases (as shown in Figure~\ref{fig:intro}e). Specific formatting rules were enforced to align model-generated responses with human annotations. During annotation, the model produced multiple candidate explanations, from which professional annotators selected and refined the most appropriate. The final annotations are integrated into a reasoning-centric framework, enabling both precise anomaly localization and high-level semantic understanding.
For privacy, all facial regions and identity-related visual attributes are blurred. Further procedural details are provided in Appendix~\ref{appendix:privacy}.

\subsection{Dataset Characteristics and Comparative Analysis}
Unlike conventional surveillance datasets, where anomalies often occupy the most salient portions of the frame, aerial-captured anomalies are typically small, scattered, and heavily context-dependent due to increased altitude, oblique angles, and limited focal length. As a result, the decisive information for anomaly detection lies not in the entire frame but in semantically rich yet spatially tiny regions, referred to as region-aware anomalies. The A2Seek dataset is specifically curated to address this challenging setting, providing a benchmark that closely mirrors real-world aerial inspection demands. 
%
\begin{table*}[!t]
\caption{Comparison of A2Seek with existing video anomaly detection datasets (\textsuperscript{\dag} denotes web-sourced datasets; \textsuperscript{\ddag} denotes simulated or virtual datasets).}
\centering
\resizebox{1.0\textwidth}{!}{
\begin{tabular}{clrrrccrcccc}
\toprule
\multirow{2}{*}{\textbf{Perspective} \vspace{-0.5em}} & \multirow{2}{*}{\textbf{Dataset} \vspace{-0.5em}} & \multicolumn{3}{c}{\textbf{Frames}} & \multirow{2}{*}{\makecell{\textbf{Scene} \\ \textbf{Count}} \vspace{-0.5em}} & \multirow{2}{*}{\makecell{\textbf{Anomaly} \\ \textbf{Types}} \vspace{-0.5em}} & \multirow{2}{*}{\textbf{Resolution} \vspace{-0.5em}} &  \multirow{2}{*}{\makecell{\textbf{Scene} \\ \textbf{Dependency}} \vspace{-0.5em}} & \multirow{2}{*}{\makecell{\textbf{Scale} \\ \textbf{Variation}} \vspace{-0.5em}} & \multirow{2}{*}{\makecell{\textbf{Reasoning} \\ \textbf{Annotation}} \vspace{-0.5em}} & \multirow{2}{*}{\makecell{\textbf{Multi-} \\ \textbf{modal}} \vspace{-0.5em}} \\
\cmidrule{3-5}
& & \textbf{Total} & \textbf{Normal} & \textbf{Abnormal} &  &  &  &  &  &  &   \\ 
 \midrule
\multirow{13}{*}{Surveillance}  & CUHK Avenue \cite{lu2013abnormal}      & 30,652         & 26,832          & 3,820             & 1              & 5              & 640$\times$360       &  \textcolor{lightgray}{\ding{55}}         & \textcolor{lightgray}{\ding{55}}         & \textcolor{lightgray}{\ding{55}}         & \textcolor{lightgray}{\ding{55}}         \\

 & ShanghaiTech \cite{luo2017revisit}    & 317,398        & 300,308         & 17,090            & 13             & 11             & 856$\times$480       & \textcolor{lightgray}{\ding{55}}         & \textcolor{lightgray}{\ding{55}}         & \textcolor{lightgray}{\ding{55}}         & \textcolor{lightgray}{\ding{55}}         \\
 
   &Street Scene \cite{ramachandra2020street}     & 203,257        & 159,341         & 43,916            & 1              & 17             & 1,280$\times$720      &  \textcolor{lightgray}{\ding{55}}         & \textcolor{lightgray}{\ding{55}}         & \textcolor{lightgray}{\ding{55}}         & \textcolor{lightgray}{\ding{55}}         \\
   
   
   & Subway \cite{adam2008robust} & 209,151 & 192,548 & 16,603 & 2 & 8 & 512$\times$384 & \textcolor{lightgray}{\ding{55}} & \textcolor{lightgray}{\ding{55}} & \textcolor{lightgray}{\ding{55}} & \textcolor{lightgray}{\ding{55}} \\
   & UBI-Fight\textsuperscript{\dag}  \cite{degardin2021iterative} & 8,530,080 & 8,287,381 & 242,699 & - & 1 & 1,280$\times$720 & \textcolor{lightgray}{\ding{55}} & \textcolor{lightgray}{\ding{55}} & \textcolor{lightgray}{\ding{55}} & \textcolor{lightgray}{\ding{55}} \\
    & LAD\textsuperscript{\dag} \cite{wan2021anomaly}       & 3,625,237 & 3,016,213 & 609,024 & - & 14 & 320$\times$240 & \textcolor{lightgray}{\ding{55}} & \textcolor{lightgray}{\ding{55}} & \textcolor{lightgray}{\ding{55}} & \textcolor{lightgray}{\ding{55}} \\
   
    &IITB Corridor \cite{chan2017anticipating}    & 483,566        & 301,999         & 181,567           & 1              & 10             & 1,920$\times$1,080     &  \textcolor{lightgray}{\ding{55}}         & \textcolor{lightgray}{\ding{55}}         & \textcolor{lightgray}{\ding{55}}         & \textcolor{lightgray}{\ding{55}}         \\
    
    
    & UCSD Ped2 \cite{li2013anomaly}        & 4,560          & 2,924           & 1,636             & 1              & 5              & 360$\times$240       & \textcolor{lightgray}{\ding{55}}         & \textcolor{lightgray}{\ding{55}}         & \textcolor{lightgray}{\ding{55}}         & \textcolor{lightgray}{\ding{55}}         \\
    
    & UMN \cite{raghavendra2006unusual}              & 7,741          & 6,165           & 1,576             & 3              & 1              & 320$\times$240       &  \textcolor{lightgray}{\ding{55}}         & \textcolor{lightgray}{\ding{55}}         & \textcolor{lightgray}{\ding{55}}         & \textcolor{lightgray}{\ding{55}}         \\

   & XD-Violence\textsuperscript{\dag} \cite{wu2020not}      &   18,697,729      &       -      &       -  &     -       &       6       &       640$\times$360       & \textcolor{lightgray}{\ding{55}}      & \textcolor{lightgray}{\ding{55}}      & \textcolor{lightgray}{\ding{55}}      & \ding{51}   \\
   & UCF-Crime\textsuperscript{\dag} \cite{yuan2024towards}        &   13,741,393      &     -      &       -       &      -      &       13       &       320$\times$240       & \textcolor{lightgray}{\ding{55}}      & \textcolor{lightgray}{\ding{55}}      & \textcolor{lightgray}{\ding{55}}      & \textcolor{lightgray}{\ding{55}}      \\
    & TAD \cite{lv2021localizing} & 540,272 & 462,578 & 77,694 & - & 7 & 1,280$\times$720  &  \textcolor{lightgray}{\ding{55}}         & \textcolor{lightgray}{\ding{55}}         & \textcolor{lightgray}{\ding{55}}         & \textcolor{lightgray}{\ding{55}}         \\
   & UBnormal\textsuperscript{\ddag} \cite{acsintoae2022ubnormal}         & 236,902        & 147,887         & 89,015            & 29             & 22             & 1,080$\times$720      &  \textcolor{lightgray}{\ding{55}}         & \textcolor{lightgray}{\ding{55}}         & \textcolor{lightgray}{\ding{55}}         & \textcolor{lightgray}{\ding{55}}         \\
   
    & NWPU Campus \cite{cao2023new}      & 1,466,073      & 1,400,807       & 65,266            & 43             & 28             & 2,048$\times$1,536       &   \textcolor{black}{\ding{51}}      & \textcolor{lightgray}{\ding{55}}        & \textcolor{lightgray}{\ding{55}}        & \textcolor{lightgray}{\ding{55}}        \\
    

    & MSAD\textsuperscript{\dag} \cite{zhu2024advancing} & 447,236     & 308,792     & 138,444     & 14     & 55   & 1,920$\times$1,080& \textcolor{lightgray}{\ding{55}}& \textcolor{lightgray}{\ding{55}}& \textcolor{lightgray}{\ding{55}} & \textcolor{lightgray}{\ding{55}} \\

\midrule
\multirow{4}{*}{Drone-view}            & Minidrone \cite{bonetto2015privacy}        & 23,295         & 14,821          & 8,474             & 1              & 10             & 1,280$\times$720      &  \textcolor{lightgray}{\ding{55}}         & \textcolor{lightgray}{\ding{55}}         & \textcolor{lightgray}{\ding{55}}         & \textcolor{lightgray}{\ding{55}}         \\

 & AU-AIR-Anomaly \cite{bozcan2021context}   & 32,823         & 30,000          & 2,823             & 1              & 8  
 & 1,920$\times$1,080     &  \textcolor{lightgray}{\ding{55}}         & \textcolor{lightgray}{\ding{55}}   & \textcolor{lightgray}{\ding{55}}         & \textcolor{black}{\ding{51}}        \\
 
 & Drone-Anomaly \cite{jin2022anomaly}    & 87,488         & 51,635          & 35,853            & 7              & 1    & 640$\times$640       &  \textcolor{lightgray}{\ding{55}}         & \textcolor{lightgray}{\ding{55}}         & \textcolor{lightgray}{\ding{55}}        & \textcolor{lightgray}{\ding{55}}        \\
 
 & UIT-ADrone \cite{tran2023uit} & 206,194   & 142,709 & 63,485 & 3 & 10 & 1,920$\times$1,080 & \textcolor{lightgray}{\ding{55}} & \textcolor{lightgray}{\ding{55}} & \textcolor{lightgray}{\ding{55}} & \textcolor{lightgray}{\ding{55}} \\
 
\rowcolor{gray!10}
 & \textbf{A2Seek (Ours)}    & \textbf{2,485,859}      & \textbf{2,087,160}       & \textbf{398,699}           & \textbf{27}             & \textbf{20}             & \textbf{3,840$\times$2,160}             &  \ding{51}    & \ding{51}      & \ding{51}    & \ding{51}       \\ 
\bottomrule
\end{tabular}
}
\vspace{-1em}
\label{tab:overall statics}
\end{table*}

Table~\ref{tab:overall statics} compares A2Seek with existing aerial and ground-based anomaly detection datasets. As the largest aerial anomaly detection dataset, A2Seek comprises over 2.4 million frames, including 398k frames containing anomalies, significantly surpassing other drone-view datasets in both scale and diversity. 
The dataset covers 10 major scene categories, such as roadways, sidewalks, playgrounds, and industrial zones. These categories are further divided into 27 specific sub-scenes, providing comprehensive coverage of complex real-world environments. \par
Beyond its scale and diversity, A2Seek is the first aerial anomaly detection dataset to provide natural language reasoning annotations with precise spatial localization labels. These annotations enable models to perform deep anomaly understanding and reasoning, bridging the gap between detection and interpretability. Furthermore, A2Seek explicitly addresses challenges unique to aerial perspectives, such as scene dependency, subtle anomalies, complex backgrounds, and scale variations, while also integrating multimodal data (RGB and infrared) to handle diverse environmental conditions.
In summary, A2Seek excels in data modalities, scene richness, and fine-grained annotations, directly addressing the two core challenges of aerial anomaly detection: spatial localization and semantic generalization. By establishing a robust data foundation and offering clear research directions, A2Seek sets a new benchmark for advancing aerial video anomaly detection and promoting deeper exploration in this domain. 
More details about our A2Seek data are reported in Appendix \ref{appendix:data_part}.

\section{Methodology}
\label{method}

Anomaly detection from aerial perspectives poses unique challenges, requiring both precise spatial localization (``Where is the anomaly?'') and comprehensive semantic reasoning (``Why is it an anomaly?''). To address these challenges, we propose A2Seek-R1, a reasoning-centric framework specifically designed for aerial anomaly understanding. The framework integrates two key stages: (1) Supervised Fine-Tuning (SFT), which leverages Graph-of-Thought (GoT) annotations to activate the model's initial reasoning capabilities by structuring reasoning into explicit semantic stages, and (2) Reinforcement Fine-Tuning (RFT), which employs Aerial Group Relative Policy Optimization (A-GRPO) to refine the model's reasoning and localization strategies through task-specific reward functions. 
Among them, the seeking reward enables the model to dynamically identify regions requiring additional high-resolution context for further analysis, while the length reward encourages concise answers in simple scenarios and allocates more reasoning effort to complex and challenging scenes.
Through A-GRPO-driven reinforcement fine-tuning, the model not only achieves accurate anomaly detection but also generates interpretable and verifiable reasoning traces, making it effective for real-world aerial applications.

\textbf{STAGE 1: SFT-based Reasoning Activation. }
The first stage aims to activate the model's initial reasoning capabilities through supervised fine-tuning (SFT) on the A2Seek dataset, specifically designed for aerial anomaly understanding. Each sample consists of an $n$-frame clip, an anomaly behavior label, and optional annotations such as different steps of Graph-of-Thought (GoT) reasoning or candidate bounding boxes. Missing fields are represented by a placeholder token {\small\texttt{<NULL>}}.
To accommodate diverse supervision, we apply a binary mask over the target sequence, activating gradients only on annotated tokens. 
The supervised loss is defined as:
\begin{equation}
\mathcal{L}_{\mathrm{SFT}} = -\,\mathbb{E}_{(x,r,b,a)\sim\mathcal{D}_{\mathrm{A2Seek}}}
\sum_{t=1}^{T}
m_t \log \pi_\theta(y_t \mid x, y_{<t}),
\end{equation}
where \(\mathcal{D}_{\mathrm{A2Seek}}\) is the data distribution, \(x\) represents the input frames, \(r\) denotes the reasoning sequence, \(b\) refers to the bounding boxes, and \(a\) is the anomaly label. The binary mask \(m_t\) indicates whether the \(t\)-th token in the target sequence \(y\) is supervised (\(m_t = 1\)) or not (\(m_t = 0\)). The model's output distribution \(\pi_\theta(y_t \mid x, y_{<t})\) represents the conditional probability of generating the \(t\)-th token \(y_t\), given the input \(x\) and the preceding tokens \(y_{<t}\). The loss is computed over all supervised tokens in the sequence.

To enhance interpretability, we introduce a Graph-of-Thought (GoT) annotation (as shown in Figure \ref{fig:intro}e) that structures reasoning into explicit semantic stages, including visual signals ({\small{\texttt{<|Trigger|>}}}), anomaly diagnosis ({\small{\texttt{<|Diagnose|>}}}), causal explanation ({\small{\texttt{<|Reasoning|>}}}), confidence reflection ({\small{\texttt{<|Reflection|>}}}), and follow-up action ({\small{\texttt{<|Seeking|>}}}). 
Among these, the {\small{\texttt{<|Seeking|>}}} tag introduces a novel mechanism that enables the model to actively identify regions requiring additional high-resolution context, rather than merely localizing anomalies. 
Specifically, the model predicts a bounding box representing a potential region of interest, rather than a specific object, to be cropped and analyzed further. This mechanism mimics the behavior of UAVs actively adjusting their viewpoints to gather more detailed information. Unlike localization, which aims to pinpoint the exact position of an anomaly, seeking emphasizes identifying broader regions that warrant further inspection, enabling the model to effectively handle complex or ambiguous scenarios.
These tags are injected into the target sequence and jointly trained under the same objective, guiding the model to generate coherent and verifiable reasoning traces.
This structured supervision enables the model to generalize across varying annotation levels, from label-only samples to full reasoning and localization instances, providing a foundation for interpretable downstream adaptation.

\textbf{STAGE 2: RL-based Reasoning Exploration. }
The second stage refines the model's reasoning and localization strategies through reinforcement fine-tuning (RFT). This stage aims at addressing the dynamic complexity of aerial anomaly detection, where environmental conditions and scene semantics vary significantly. Anomaly understanding could be formulated as a sequential decision-making problem and optimize the model using reinforcement learning. 
Following \cite{swamy2025all}, we define the model's output as a policy \(\pi_\theta(y|x)\), which represents the model's output distribution parameterized by \(\theta\).
The objective is to maximize the expected reward:
\begin{equation}
J(\theta) = \mathbb{E}_{y \sim \pi_\theta(y|x)} \left[ R(x, y) \right],
\end{equation}
where \(R(x, y)\) measures the quality of the output \(y\) for the given input \(x\). 
The reward \(R(x, y)\) is composed of multiple components, including format, accuracy, localization, seeking, and length rewards.
Format and accuracy rewards constrain the model to produce outputs in the correct format and focus on prediction accuracy, while the localization reward evaluates the model's judgment of anomaly evidence. The seeking and length rewards hierarchically guide the model to achieve dynamic reasoning paths.
More detailed reward function designs are provided in Appendix~\ref{appendix:reward_design}.

Building on this, we formulate reinforcement fine-tuning as a reward-weighted log-likelihood maximization problem with prior regularization. Specifically, let \(\mathcal{X}\) denote the input space, which consists of sequences of aerial video frames, and \(\mathcal{Y}\) denote the output space, which includes structured reasoning traces and anomaly prediction. A policy \(\pi \in \Pi\) maps inputs \(x \in \mathcal{X}\) to a distribution over outputs \(y \in \mathcal{Y}\). The reinforcement fine-tuning objective is defined as:
\begin{equation}
\pi^\star = \argmax_{\pi \in \Pi} \mathbb{E}_{x \sim \mathcal{D}_{\mathrm{A2Seek}}} \mathbb{E}_{y \sim \pi(y|x)} [R(x, y)] - \beta \cdot D_{\text{KL}}(\pi(y|x) \,\|\, \pi_{\text{ref}}(y|x)),
\end{equation}
where \(\pi_{\text{ref}}\) is a reference policy. The KL divergence regularizes \(\pi\) to stay close to \(\pi_{\text{ref}}\), with \(\beta\) controlling the trade-off between reward maximization and regularization.
To implement this efficiently, we adopt Aerial Group Relative Policy Optimization (A-GRPO). 
Unlike existing reinforcement learning approaches \cite{chen2025r1v, liu2025visual}, our A-GRPO algorithm explicitly incorporates anomaly location and region seeking stages, making it particularly effective for aerial anomaly understanding.
For each input \(x\), \(K\) candidates \(\{y^{(1)}, \dots, y^{(K)}\}\) are sampled from a policy group, and reward rankings are computed. The best-performing candidate is used to update the main policy via gradient ascent:
\begin{equation}
\nabla_\theta J(\theta) \approx \nabla_\theta \log \pi_\theta(y^{(k^*)}|x) \cdot \left( R(x, y^{(k^*)}) - b(x) \right),
\end{equation}
where \(k^* = \argmax_k R(x, y^{(k)})\), and \(b(x)\) is a baseline, defined as the mean reward of the group, to reduce variance during optimization. 
This reflection-aware RFT stage leverages A2Seek's comprehensive annotations to refine reasoning behaviors, enabling the model to generalize to unseen environments and adaptively revise predictions.
More theoretical analysis on the effectiveness of our GoT data and seeking mechanism is discussed in the Appendix \ref{appendix:discuss_effectiveness}.

\section{Experiment}

\subsection{Implementation Details}

We employ Qwen2.5-VL-3B-Instruct \cite{bai2025qwen2} as the base model due to its strong performance in vision-language understanding. We employed
LoRA \cite{hu2022lora} to adjust all fully connected layers within the model, and set a learning rate of $1\mathrm{e}{-5}$, using 4 epochs for SFT and 1 epoch for RL. Input resolution is set to $896 \times 448$, with $4$ frames per sequence. Batch size is 1, and gradient accumulation is performed over 16 steps. A cosine learning rate scheduler with a warm-up phase (5\%) is used. All experiments are conducted in PyTorch on a platform with an Intel Xeon Platinum 8350 CPU, four NVIDIA A100 GPUs, and 1,024 GB of memory.



\begin{table*}[thp]
\centering
\caption{Performance comparison across different scenarios on the A2Seek benchmark.}
\begin{adjustbox}{max width=0.9\textwidth}
\begin{tabular}{l|C{1.2cm}C{1.2cm}C{1.2cm}C{1.2cm}C{1.2cm}C{1.2cm}C{1.2cm}C{1.2cm}C{1.2cm}C{1.2cm}|C{1.5cm}}
\toprule
\multirow{2}{*}{\textbf{Method} \vspace{-0.5em}} & \multicolumn{10}{c|}{\textbf{Scene Indices}} & \multirow{2}{*}{\textbf{Average} \vspace{-0.5em}} \\
\cmidrule(lr){2-11}
 & S00 & S01 & S02 & S03 & S04 & S05 & S06 & S07 & S08 & S09 &  \\
\midrule
SSRL \cite{li2022scale}           & 11.50                & 10.20                & \textcolor{white}{0}8.30                 & \textcolor{white}{0}9.40                 & 20.10                & 12.70                & \textcolor{white}{0}5.80                 & 14.20                & 16.80                & 11.00                & 12.00                \\
HSTforU \cite{le2025hstforu} & 47.41 &	50.18 &	43.52 &	30.03	& 29.68	& 25.01 &	{\textcolor{white}{0}}9.70 &	37.28 &	48.72 &	33.69 &  47.66\\
ANDT \cite{jin2022anomaly} & 10.89	& 39.00	& 48.82	& 25.21	& 20.20	& 19.94 &	30.18 &	37.68 &	49.99 &	28.63 & 40.42 \\
AnomalyRuler \cite{yang2024anomalyruler}        & 17.40                & 15.00                & \textcolor{white}{0}9.80                 & 22.50                & 25.60                & 14.90                & \textcolor{white}{0}6.00                 & 19.30                & 21.70                & 18.80                & 17.10                \\
LAVAD \cite{zanella2024harnessing}               & 12.80                & 15.20                & 11.00                & 19.50                & 24.70                & 20.50                & \textcolor{white}{0}4.30                 & 19.10                & 22.10                & 14.90                & 16.41                \\
Holmes-VAU  \cite{zhang2024holmes}         & 17.00                & 18.50                & 16.00                & 23.40                & 25.00                & 20.50                & 13.34                & 21.12                & 27.20                & 17.60                & 19.97                \\
LLavaVideo \cite{liu2024visual}         & 12.00                & 10.00                & 16.00                & 15.00                & 24.00                & 13.00                & \textcolor{white}{0}4.00                 & 23.00                & 12.00                & 22.00                & 15.10                \\
InternVL-3.0 \cite{chen2024internvl} & 35.62 & 49.75 & 48.21 & 47.08 & 27.43 & 23.94 & 72.40 & 29.64 & 36.99 & 42.33 & 41.34 \\

\midrule
Zero-shot \cite{bai2025qwen2}           & 44.83                & 44.92                & 35.15                & 23.88                & 22.42                & 19.16                & 62.80                & 30.05                & 29.44                & 27.62                & 34.03                \\
RL-Zero              & \textcolor{white}{0}0.56                 & 14.32                & \textcolor{white}{0}7.59                 & \textcolor{white}{0}2.06                 & 29.10                & \textcolor{white}{0}4.94                 & \textcolor{white}{0}1.60                 & 16.07                & 18.84                & 29.10                & 12.42                \\
ANS-SFT              & \textcolor{white}{0}7.87                 & 32.10                & 45.89                & 25.12                & 32.63                & 16.79                & 44.00                & 34.29                & 47.95                & 14.29                & 30.09                \\
CoT-SFT              & 42.81                & 51.36                & 42.41                & 62.20                & 23.47                & 25.02                & 73.60                & 25.36                & 24.66                & 39.68                & 41.06                \\
GoT-SFT             & 49.44                & 54.32                & 54.02                & 64.96                & 32.34                & 23.20                & 63.20                & 26.43                & 54.45                & 41.80                & 46.42                \\
\rowcolor{gray!10}
A2Seek-R1            & 51.12                & 56.54                & 55.36                & 75.60                & 40.84                & 26.86                & 81.60                & 38.21                & 61.30                & 43.92                & 53.14       \\
\bottomrule
\end{tabular}
\end{adjustbox}
\label{tab:performance_comparison}
\vspace{-1em}
\end{table*}

\subsection{Overall Performance}
To highlight the contributions of our dataset and method, we design the following experimental settings: (1) Zero-Shot, serving as a baseline without any fine-tuning; (2) RL-Zero, which applies GRPO with basic format and accuracy rewards; (3) ANS-SFT, utilizing anomaly labels for supervised fine-tuning; (4) CoT-SFT, incorporating chain-of-thought to guide inference; and (5) GoT-SFT, leveraging our graph-of-thought reasoning data to enable structured anomaly understanding.
We evaluate all methods using category-level average precision ($AP_c$) for anomaly detection and mean Intersection over Union ($mIoU$) for localization, providing a balanced assessment of classification and grounding performance.




\textbf{Scene-wise Performance across Environments.} 
As shown in Table~\ref{tab:performance_comparison}, the proposed A2Seek-R1 consistently outperforms other methods in most cases, demonstrating its robustness and adaptability. Notably, in the \emph{Playground} (S03) and \emph{Rooftop} (S06) scenes, our method achieves the highest scores of 75.6\% and 81.6\%, respectively, significantly surpassing the second-best method. On average, A2Seek-R1 achieves a score of 53.14\%, which is 19.11\% higher than the baseline and 6.52\% higher than GoT-SFT. These results highlight the effectiveness of our proposed method in aerial anomaly understanding and its strong generalization capability across diverse environments.

\begin{table}[htbp]
\centering
\vspace{-1em}
\caption{Evaluation of anomaly detection and localization.} 
\begin{subtable}[t]{0.51\linewidth} 
\centering
\caption{Average precision and language metrics}
\adjustbox{max width=\linewidth}{ 
\begin{tabular}{l|cccccc}
\toprule
\textbf{Method} & \textbf{$AP_c$} & \textbf{$BLEU$} & \textbf{$METEOR$} & \textbf{$ROUGE$} & \textbf{$CIDEr$} & $Samples/s$ \\
\midrule
Zero-Shot & 34.03 & 0.2835 & 0.2145 & 0.3263 &  0.8147 & 0.6097\\
ANS-SFT & 31.10 & 0.3821 & 0.3452 & 0.4379 & 1.0462 & 0.9174 \\
GOT-SFT & 46.42 & 0.4478 & 0.3325 & 0.4439 & 1.1528 & 0.2890 \\
\rowcolor{gray!10}
A2Seek-R1 & 53.14 & 0.4564 & 0.3543 & 0.4882 & 1.2989 & 0.3267 \\
\bottomrule
\end{tabular}
}
\label{tab:lang_iou_a}
\end{subtable}
\hfill 
\begin{subtable}[t]{0.44\linewidth} 
\centering
\caption{Localization performance} 
\adjustbox{max width=\linewidth}{ 
\begin{tabular}{l|ccccccc}
\toprule
\textbf{Method} & \textbf{$mIoU$} & \textbf{$AP_{0.00}$} & \textbf{$AP_{0.25}$} & \textbf{$AP_{0.50}$} & \textbf{${AP_{0.75}}$} & ${AP_{0.90}}$ \\
\midrule
Zero-Shot & \textcolor{white}{0}3.50 &	72.40 &	\textcolor{white}{0}3.94 &	\textcolor{white}{0}1.81 &	\textcolor{white}{0}0.63 &	0.00  \\
ANS-SFT & 17.05 & 50.70 &	40.01	& 31.94	& 15.74 &	0.70 \\
GOT-SFT & 20.81 & 43.85 &	35.88	& 27.60	& 16.01 &	1.03 \\
\rowcolor{gray!10}
A2Seek-R1 & 26.03 &	53.31 &	45.43	& 35.33 &	20.11 &	4.34 \\
\bottomrule
\end{tabular}
}
\label{tab:lang_iou_b}
\end{subtable}
\end{table}

\textbf{Language Semantic Evaluation.} 
To evaluate the ability of models to handle semantic ambiguity and similarity in aerial anomaly understanding, we adopt language-based metrics such as BLEU~\cite{papineni2002bleu}, METEOR~\cite{banerjee2005meteor}, ROUGE~\cite{lin2004rouge}, and CIDEr~\cite{vedantam2015cider}. These metrics capture both lexical and structural alignment between predicted and ground-truth descriptions. 
As shown in Table \ref{tab:lang_iou_a}, A2Seek-R1 achieves the highest scores across all metrics, with BLEU of 0.4564, METEOR of 0.3543, ROUGE of 0.4882, and CIDEr of 1.2989. 
 The throughput is 0.3267 samples per second, which is lower than ANS-SFT (0.9174) and Zero-Shot (0.6097) because our method allocates a larger token budget at test time for reasoning, leading to better anomaly understanding. 
These results demonstrate its ability to generate accurate and semantically meaningful descriptions even under ambiguous scenarios. By leveraging visual evidence during reasoning, A2Seek-R1 ensures that its language-based predictions are contextually grounded, effectively bridging the gap between semantic understanding and visual perception. 

\textbf{Supporting Region Grounding Accuracy. } 
To further evaluate whether the model relies on visual information to understand anomalies, we assess its localization performance using both mIoU and AP under varying IoU thresholds. 
As shown in Table~\ref{tab:lang_iou_b}, A2Seek-R1 achieves the highest mIoU score of 26.03\%, outperforming Zero-Shot and GoT-SFT by 22.53\% and 5.22\%, respectively. Notably, although Zero-Shot and ANS-SFT achieve relatively high AP values in $AP_{0.00}$, this is mainly attributed to their tendency to predict overly broad and vague regions, rather than accurately indicating the anomalous areas. This limitation is further reflected in their significantly degraded performance (nearly zero in $AP_{0.90}$) at higher IoU thresholds, indicating a tendency toward random guessing.
These results confirm that the spatial reasoning mechanism introduced in our approach enhances the model's ability to precisely localize subtle and ambiguous anomalies, thereby improving both detection accuracy and interpretability for real-world aerial anomaly understanding.

\begin{figure*}[htbp]
  \centering
  \begin{subfigure}[t]{0.23\linewidth}
    \centering
    \includegraphics[width=\linewidth]{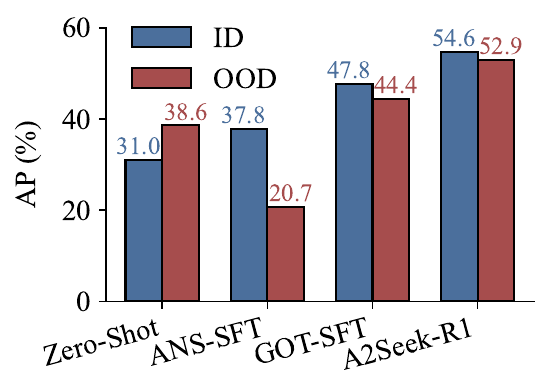}
    \caption{Overall accuracy}
  \label{fig:5_domain_a}
  \end{subfigure}
  \hfill
  \begin{subfigure}[t]{0.35\linewidth}
    \centering
    \includegraphics[width=\linewidth]{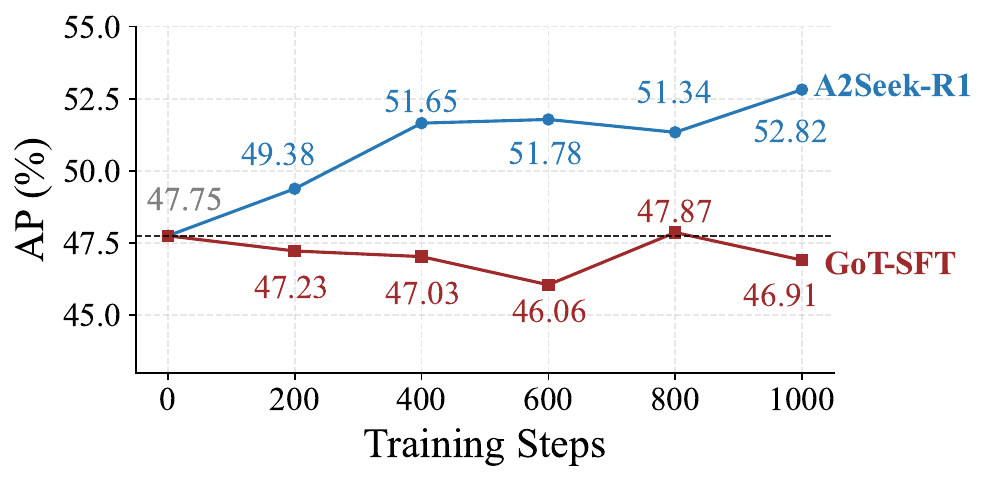}
    \caption{Data efficiency}
  \label{fig:5_domain_b}
  \end{subfigure}
  \hfill
  \begin{subfigure}[t]{0.35\linewidth}
    \centering
    \includegraphics[width=\linewidth]{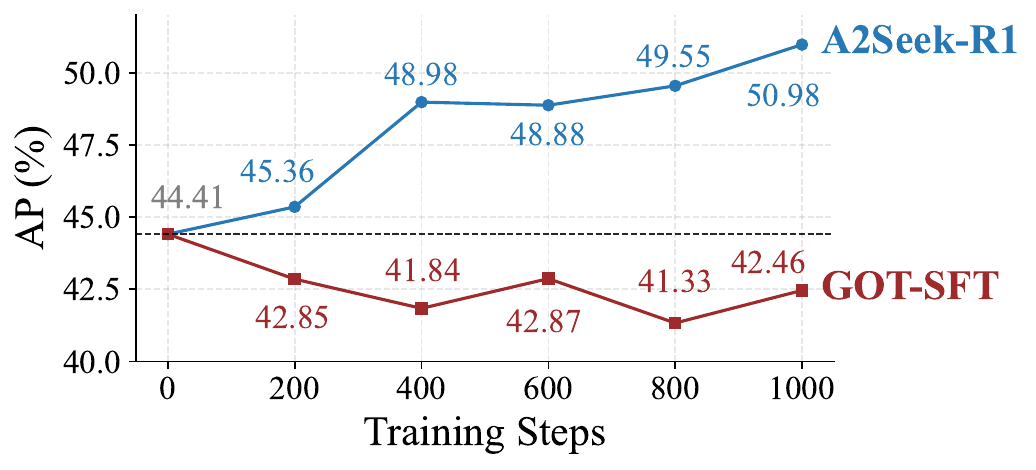}
    \caption{OOD generalization}
  \label{fig:5_domain_c}
  \end{subfigure}
  \caption{Performance comparison of different settings on A2Seek benchmark.} 
  \vspace{-1em}
  \label{fig:5_domain}
\end{figure*}

\textbf{Data Efficiency and Generalization.} The in-domain (ID) data in A2Seek includes scenarios present in the training set, while the out-of-domain (OOD) data comprises unseen scenarios, specifically designed to assess the model's generalization capability in novel and challenging settings. 
As shown in Figure \ref{fig:5_domain}, A2Seek-R1 demonstrates both high data efficiency on in-domain data and strong generalization to out-of-domain scenarios. On ID data (Figure \ref{fig:5_domain_b}), A2Seek-R1 achieves consistent performance improvements throughout training, whereas GoT-SFT shows stagnation or minor fluctuations. Notably, on OOD data (Figure \ref{fig:5_domain_c}), A2Seek-R1 continues to gain accuracy as training progresses, reaching 48.98\% at 400 steps, a relative improvement of 4.57\%. In contrast, GoT-SFT degrades from 44.41\% to 41.84\% over the same period. This observation suggests that A2Seek-R1 not only leverages training data more effectively but also exhibits better robustness under domain shifts.



\begin{table*}[hbtp]
\centering
\caption{Ablation studies of the A2Seek-R1 framework.}
\setlength{\tabcolsep}{4pt} 
\adjustbox{max width=0.75\linewidth}{
\begin{tabular}{c | C{1.3cm} C{1.3cm} C{1.3cm} | C{2.6cm} C{2.6cm} C{2.6cm} | c c}
\toprule
\multirow{2}{*}{\textbf{Idx} \vspace{-0.5em}} & \multicolumn{3}{c|}{\textbf{Supervised Fine-Tuning}} & \multicolumn{3}{c|}{\textbf{Reinforcement Learning}} & \multicolumn{2}{c}{\textbf{Metric}} \\
\cmidrule(lr){2-4} \cmidrule(lr){5-7} \cmidrule(lr){8-9}
 & ANS   & CoT         & GoT        & Accuracy Reward   & Location Reward        & Seeking Reward   & $AP_c$           & $mIoU$            \\
\midrule
0            & \textcolor{darkgray}{\ding{51}}      &    &   &       &   &   & 31.10          & 12.13          \\
1            & \textcolor{darkgray}{\ding{51}}      & \textcolor{darkgray}{\ding{51}}            &            &       &            &            & 38.24          & 17.05          \\
2            & \textcolor{darkgray}{\ding{51}}      &             & \textcolor{darkgray}{\ding{51}}           &       &            &            & 46.42          & 20.81          \\
3 & & & & \textcolor{darkgray}{\ding{51}} & \textcolor{darkgray}{\ding{51}} &  & 12.42 & \textcolor{white}{0}9.66 \\
4            & \textcolor{darkgray}{\ding{51}}      &             & \textcolor{darkgray}{\ding{51}}           & \textcolor{darkgray}{\ding{51}}      &            &            & 52.82          & 18.77          \\
5   & \textcolor{darkgray}{\ding{51}}      &             & \textcolor{darkgray}{\ding{51}}           & \textcolor{darkgray}{\ding{51}}      & \textcolor{darkgray}{\ding{51}}           &            & 51.78          & 24.03          \\
\rowcolor{gray!10}
6            & \textcolor{darkgray}{\ding{51}}      &             & \textcolor{darkgray}{\ding{51}}           & \textcolor{darkgray}{\ding{51}}      & \textcolor{darkgray}{\ding{51}}           & \textcolor{darkgray}{\ding{51}}           & 53.14          & 26.03         \\
\bottomrule
\end{tabular}
}
\vspace{-1em}
\label{tab:ablation}
\end{table*}

\subsection{Ablation Studies}

Table~\ref{tab:ablation} presents the results of ablation studies on the A2Seek dataset, analyzing the contributions of supervised fine-tuning (SFT) strategies and reinforcement learning (RL) reward components.

\textbf{Impact of Supervised Fine-Tuning on A2Seek.}
Table~\ref{tab:ablation} (Rows 0-2) evaluate the effect of different SFT strategies: ANS, CoT, and GoT. Using only ANS (Row 0) achieves a baseline $AP_c$ of 31.10\% and $mIoU$ of 12.13\%. Incorporating CoT (Row 1) improves $AP_c$ to 38.24\% and $mIoU$ to 17.05\%, demonstrating the benefit of chain-of-thought reasoning. Replacing CoT with GoT (Row 2) further boosts $AP_c$ to 46.42\% and $mIoU$ to 20.81\%, highlighting the effectiveness of goal-oriented reasoning in establishing a stronger foundation for anomaly detection and understanding.

\textbf{Impact of Reinforcement Learning Rewards.}
Rows 3-6 progressively incorporate RL reward components, including accuracy, location, and seeking rewards. Using only accuracy and location rewards (Row 5) achieves $AP_c$ of 51.78\% and $mIoU$ of 24.03\%. Adding the seeking reward (Row 6) further improves $AP_c$ to 53.14\% and $mIoU$ to 26.03\%, achieving the best overall performance. These results demonstrate that the synergistic integration of all three reward components is critical for enhancing both detection accuracy and localization precision.

\begin{table}[ht]
\centering
\vspace{-1em}
\caption{Effectiveness of A2Seek-R1 on different base models.}
\vspace{2mm}
\adjustbox{max width=0.55\linewidth}{
\begin{tabular}{lcccc}
\toprule
\textbf{Method} & \textbf{$AP_c(\%)$} & \textbf{$\Delta$$AP(\%)$} & \textbf{$mIoU(\%)$} & \textbf{$\Delta$$mIoU(\%)$} \\
\midrule
QwenVL-2.5-7B-Instruct & 44.65 & -- & 08.24 & -- \\
\rowcolor{gray!10}
\ \ \textit{+ A2Seek-R1} & 57.07 & +12.42 & 29.71 & +21.47 \\
InternVL-3.0-2B-Instruct & 20.03 & -- & 00.10 & -- \\
\rowcolor{gray!10}
\ \ \textit{+ A2Seek-R1} & 48.76 & +28.73 & 18.89 & +18.79 \\
\bottomrule
\end{tabular}
}
\label{tab:a2seek_base_models}
\end{table}

\textbf{Impact of Foundation Models.} As shown in Table~\ref{tab:a2seek_base_models}, we evaluate the generalization of A2Seek-R1 on two additional vision-language models beyond the base model, \textit{i.e.}, QwenVL-2.5-3B-Instruct~\cite{bai2025qwen2}. On QwenVL-2.5-7B-Instruct \cite{bai2025qwen2}, it improves AP by 12.42\% and mIoU by 21.47\%, indicating effective scaling with capacity. On InternVL-3.0-2B-Instruct \cite{chen2024internvl}, it increases AP from 20.03\% to 48.76\% and mIoU from 0.10\% to 18.89\%. These results demonstrate consistent gains across model sizes and architectures, supporting strong generalization and transferability.

\begin{figure*}[t!]
  \centering
    \includegraphics[width=\linewidth]{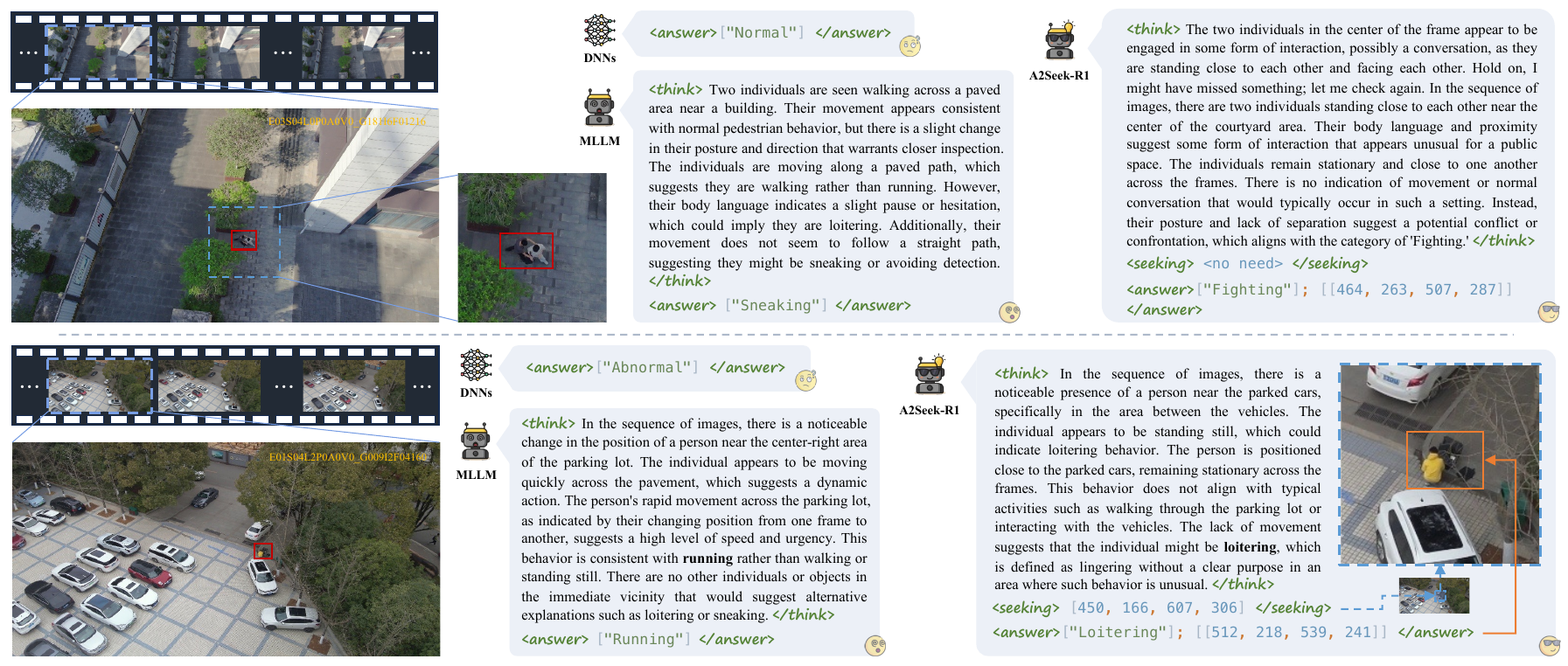}
    \vspace{-2mm}
   \caption{Qualitative results of A2Seek-R1. Beyond predicting anomaly categories, our method provides reasoning traces and accurately localizes the key regions that support its judgment.}
    \vspace{-4mm}
   \label{fig:4_vis}
\end{figure*}

\subsection{Qualitative Visualization}

As illustrated in Figure~\ref{fig:4_vis}, traditional deep neural networks (DNNs) provide only binary classifications (normal or abnormal) without explanations or spatial context. Multimodal large language models (MLLMs) improve upon this by generating textual descriptions to explain anomalies but lack the ability to localize specific regions, limiting their utility in complex aerial scenarios.
In contrast, A2Seek-R1 combines detailed reasoning with precise spatial localization. By reflecting on individuals' body language and proximity, the model iteratively refines its understanding, identifying interactions such as ``Fighting'' and providing bounding box coordinates to substantiate its reasoning. Additionally, A2Seek-R1 mimics UAV flight behavior by dynamically focusing on regions of interest, enabling adaptive refinement of predictions and enhancing both detection accuracy and localization precision.

\section{Conclusions}
\label{discussion}

%
This work introduces A2Seek, a large-scale, reasoning-centric dataset for aerial anomaly understanding. Built upon diverse aerial scenarios, A2Seek is meticulously annotated with fine-grained anomaly labels, spatiotemporal bounding boxes, and structured reasoning graphs. These annotations enable models to not only detect anomalies but also provide interpretable explanations for ``why'' and ``where'' anomalies occur. Building on this benchmark, we introduce A2Seek-R1, a novel multi-stage reinforcement fine-tuning framework that integrates graph-of-thought (GoT)-guided supervised fine-tuning, and an improved A-GRPO algorithm with tailored reward functions, including a seeking mechanism that simulates UAV flight behavior. This framework achieves state-of-the-art performance in both accuracy and interpretability, excelling in complex anomaly scenarios and demonstrating strong cross-domain generalization capabilities.

\textbf{Limitations and Societal Impact. } Despite strong performance, the complexity of reasoning chains and strategy learning limits inference speed and deployment efficiency. 
The framework mainly focuses on spatial-level anomaly perception, whereas the understanding and modeling of long-duration temporal anomalies are still limited, which we regard as an important avenue for future exploration. 
Additionally, given its potential applications in sensitive areas like public safety, privacy protection and algorithmic fairness must be prioritized to avoid societal risks from misjudgments. 


\paragraph{Acknowledgments and Disclosure of Funding.} This work was supported in part by the Science and Technology Innovation Key R\&D Program of Chongqing under Grant No. CSTB2023TIAD-STX0016, in part by the National Natural Science Foundation of China under Grants No. 62472060, U23A20318, and 62221005, in part by the Natural Science Foundation of Chongqing under Grants No. CSTB2024NSCQ-QCXMX0060 and CSTB2023NSCQ-LZX0061, in part by the Science and Technology Research Program of Chongqing Municipal Education Commission under Grant No. KJZD-K202300604, in part by the Chongqing Doctoral Student Innovation Talent Project under Grant No. CYB240241, in part by the Chongqing University of Posts and Telecommunications Ph.D. Innovative Talents Project under Grant No. BYJS202404, and in part by the Chongqing Institute for Brain and Intelligence.


\newpage

{
    \small
    \bibliographystyle{plain}
    \bibliography{bib/main}
}

\newpage

\appendix

\section{Data Collection, Annotation and Statistics}
\label{appendix:data_part}

\subsection{Dataset Acquisition Protocol}

\begin{figure*}[hbtp]
  \centering
   \includegraphics[width=\linewidth]{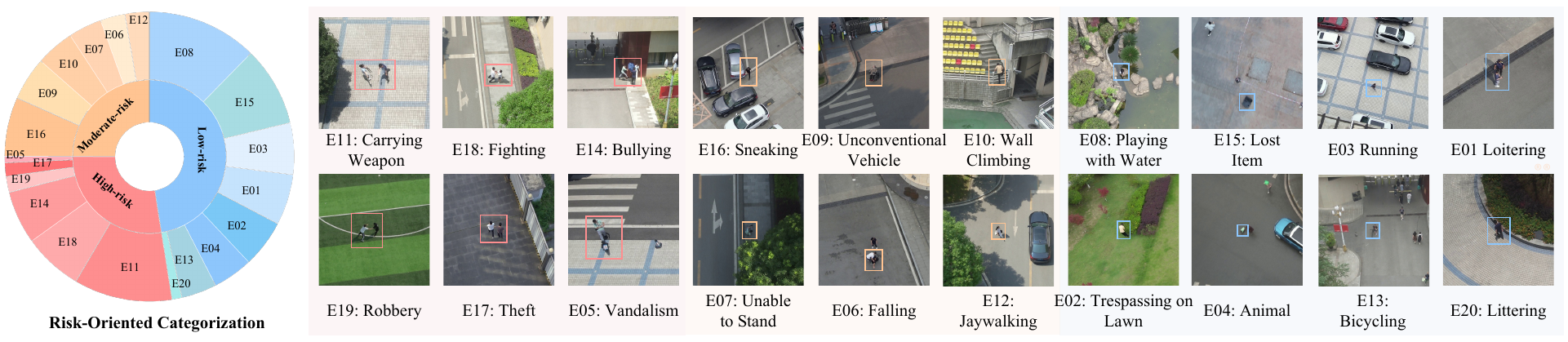}
   \caption{Representative anomaly types in the A2Seek dataset. Our dataset covers a broad spectrum of anomalous behaviors across different risk levels, highlighting the diversity and complexity of aerial anomaly detection.}
    \vspace{-2mm}
   \label{fig:anomly_type}
\end{figure*}
\begin{figure}[htbp]
  \centering
  \includegraphics[width=\textwidth]{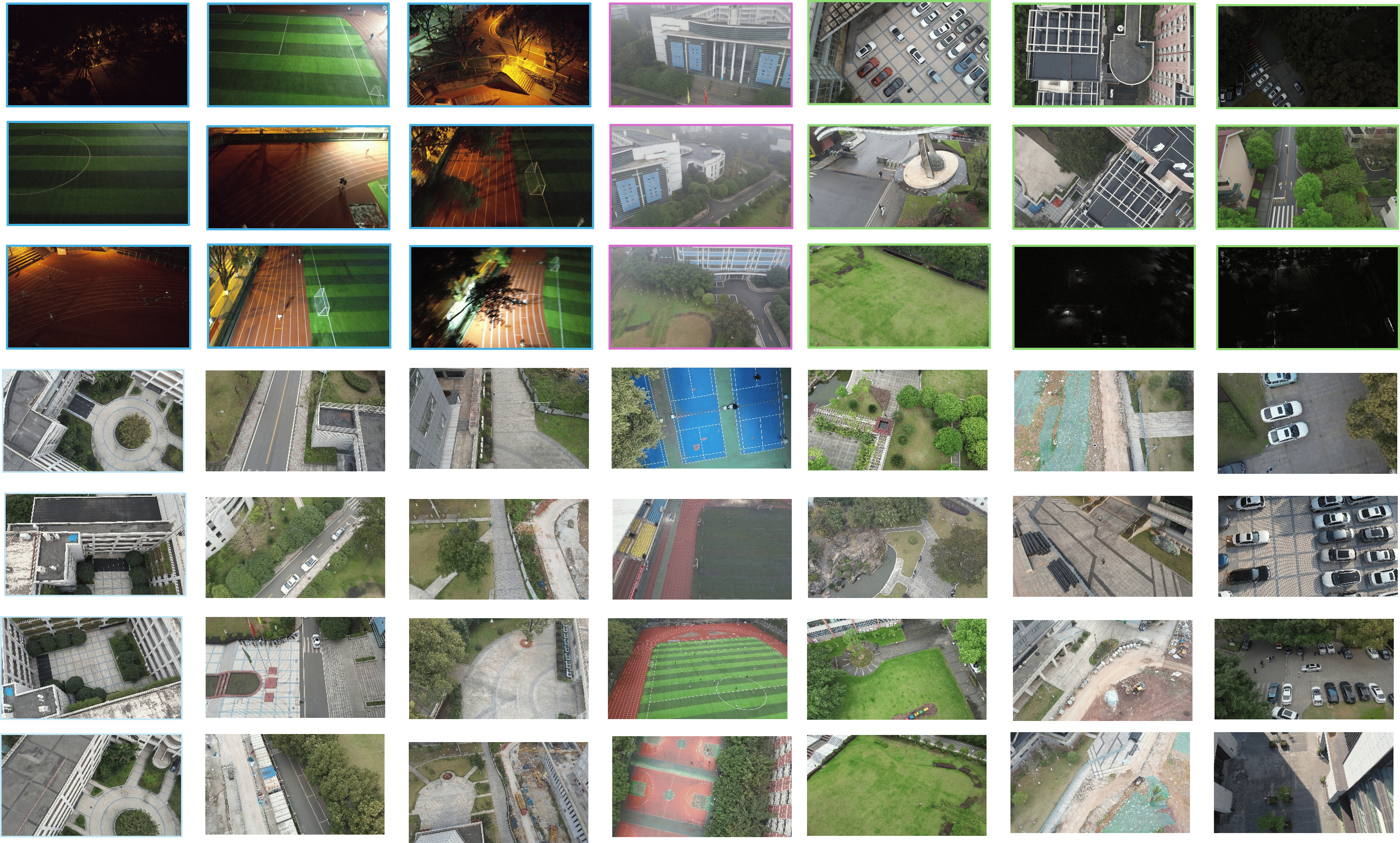}
  \caption{Examples of scenes from the A2Seek dataset.}
  \label{fig:rare_weather}
\end{figure}

To address the challenges of spatial localization and scene generalization, the data collection process was carefully designed to maximize diversity and realism. The A2Seek dataset was collected using a DJI M30T drone equipped with wide-angle, telephoto, and infrared cameras. The drone operated at varying altitudes (10 to 60 meters) and speeds (0 to 20 m/s) to capture a wide range of scale variations for anomalous objects. Flight trajectories included hovering, linear cruising, curved circling, and area scanning, reflecting dynamic viewpoint changes. 
The dataset comprises 542 untrimmed 4K videos recorded at 30 FPS across 10 campus environments, subdivided into 27 sub-scenes. Each video has an average duration of 153.74 seconds, with a total duration exceeding 23 hours. Among these, 19.3 hours feature normal behaviors, while 3.79 hours capture diverse and complex anomalies, including parallel, sequential, occluded, and scene-dependent events. As shown in Figure~\ref{fig:anomly_type}, our dataset includes over 20 anomaly categories, such as falling, fighting, playing with water, and jaywalking. Spanning nearly a year of recording, it covers various seasons, times of day (daytime, nighttime, twilight), and environmental conditions, including clear, cloudy, foggy, and rainy weather (as shown in Figure~\ref{fig:rare_weather}). This diversity provides a robust foundation for developing and evaluating aerial anomaly detection models. Table~\ref{tab:code_definitions} presents a unified overview of all categorical codes used in this paper. The \emph{Type} column groups related codes, the \emph{Code} column lists shorthand labels, and the \emph{Definition} column provides detailed descriptions.

Our collection workflow for A2Seek streamlines a campus drone survey into a single, reproducible loop while preserving fine-grained control:
\begin{itemize}
\item \textbf{Scene \& Time:} ten functional zones, further divided into 27 sub-scenes, are revisited across four seasons, three day parts (twilight: 06:00 to 08:00 or 16:00 to 19:00, daylight: 10:00 to 16:00, night: 19:00 to 22:00) and five weather classes, producing initial multimodal videos.
\item \textbf{Airframe \& Optics:} a DJI M30T (1-inch CMOS, 20 MP, 3,840$\times$2,160@30 fps) plus a co-aligned 640$\times$512 thermal sensor for nocturnal sorties; the on-board GNSS–IMU logs pose at 50 Hz.
\item \textbf{Flight Envelope:} grid missions with 80\% side and 70\% forward overlap (for routine coverage), dynamically switchable to follow or orbit mode for rare anomalies, at altitudes of 10--60\,m above ground level ($\le$\,30\,m at night) and ground speeds of 0--15\,m/s.
\item \textbf{Data Handling:} RGB, thermal and telemetry streams are recorded to a 512 GB UHS-II SD card, then mirrored to an offline workstation upon landing; filenames are encoded with UTC timestamps and scene IDs for instant indexing.
\item \textbf{Quality Gate:} all sorties are checked daily by two senior annotators for focus, exposure and occlusion. Flights flagged for privacy concerns or other disqualifying conditions are scheduled to be re-flown.
\end{itemize}
This lean yet rigorous protocol provides high-resolution, context-rich footage tailored to drone-view anomaly detection while keeping field operations swift and fully repeatable.

\label{fig:label_process}
\begin{figure*}[t]
    \centering
    \includegraphics[width=\textwidth]{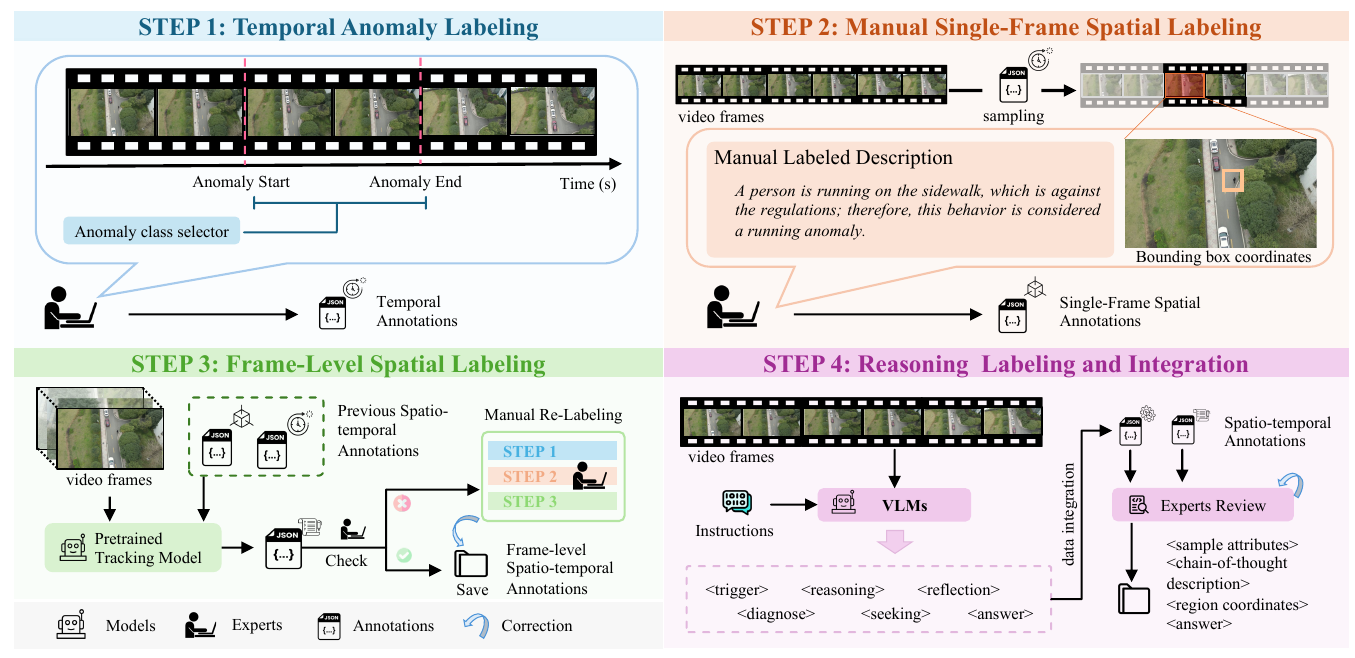}%
    \caption{Four–stage annotation workflow. 
    \emph{Step 1} (blue) Temporal annotators use our in-house GUI to mark the start/end frames and class of every anomalous episode, exporting a JSON timeline.
    \emph{Step 2} (salmon) For the first frame of each event, experts draw a bounding box around the anomalous region and supply a natural-language description, yielding single-frame spatial seeds.
    \emph{Step 3} (green) A pretrained tracker propagates each seed through the clip to form full-length trajectories; an automated checker screens the results, flags uncertain cases for human correction, and funnels all approved tracks into the spatial-label repository.
    \emph{Step 4} (violet) Vision-language models (VLMs) ingest the temporal tags, spatial tracks, and human captions; via chain-of-thought reasoning, they merge these cues into consolidated frame-level annotations, producing the final label set.}

    \label{fig:label_pipeline}
\end{figure*}

%


\begin{table}[htbp]
  \centering
  \caption{Unified code definitions for scenes, actions, parameters, weather conditions, and risk levels.}
  \vspace{2mm}
  \label{tab:code_definitions}
  \begin{adjustbox}{max width=\textwidth}
  \begin{tabular}{>{\centering\arraybackslash}r ll}
    \toprule
    \textbf{Type} & \textbf{Code} & \textbf{Definition} \\
    \midrule
    \multirow{10}{*}{Scene} & S00 & Miscellaneous: other or unspecified scenes \\
    & S01 & Roadway: street and vehicular paths \\
    & S02 & Sidewalk: pedestrian walkways along roads \\
    & S03 & Playground: open recreational areas \\
    & S04 & Open Area: unstructured outdoor space \\
    & S05 & Park: landscaped public green spaces \\
    & S06 & Rooftop: building rooftop areas \\
    & S07 & Entrance: school gate or entryway region \\
    & S08 & Wall Zone: perimeter walls and boundaries \\
    & S09 & Academic Building: building facade and courtyard \\
    \hline
    \addlinespace
    \multirow{20}{*}{Action} & E00 & Normal: no anomalous activity \\
    & E01 & Loitering: prolonged stationary presence near entrances or perimeter areas \\
    & E02 & Trespassing on Lawn: entering grass areas illegally \\
    & E03 & Running: fast movement on foot in restricted or non-sport areas \\
    & E04 & Animal: appearance of animals within the scene \\
    & E05 & Vandalism: intentional damage to property \\
    & E06 & Falling: loss of balance leading to a fall \\
    & E07 & Unable to Stand: person unable to recover after a fall \\
    & E08 & Playing with Water: interacting with fountains or water bodies \\
    & E09 & Unconventional Vehicle: use of nonstandard or unauthorized vehicles \\
    & E10 & Wall Climbing: scaling vertical surfaces \\
    & E11 & Carrying Weapon: person holding or drawing a weapon \\
    & E12 & Jaywalking: crossing road outside crosswalk \\
    & E13 & Bicycling: riding a bicycle in pedestrian or restricted areas \\
    & E14 & Bullying: aggressive behavior towards others \\
    & E15 & Lost Item: dropping or leaving behind personal belongings \\
    & E16 & Sneaking: moving stealthily or furtively \\
    & E17 & Theft: unauthorized taking of property \\
    & E18 & Fighting: physical aggression between persons \\
    & E19 & Robbery: theft involving force or threat \\
    & E20 & Littering: discarding waste improperly in public areas \\
    \hline
    \addlinespace
    \multirow{2}{*}{Height} & H0 & 10--20\,m (Low altitude flight) \\
    & H1 & 20--50\,m (High altitude flight) \\
    \hline
    \addlinespace
    \multirow{3}{*}{Velocity} & M0 & 0--3\,m/s (Low speed) \\
    & M1 & 3--10\,m/s (Moderate speed) \\
    & M2 & 10--20\,m/s (High speed) \\
    \hline
    \addlinespace
    \multirow{3}{*}{Time of Day} & L0 & 10:00--16:00 (Daytime) \\
    & L1 & 08:00--10:00, 16:00--18:00 (Morning/Evening) \\
    & L2 & 18:00--22:00 (Evening/Night) \\
    \hline
    \addlinespace
    \multirow{6}{*}{Weather} & W0 & Clear: no precipitation or clouds \\
    & W1 & Cloudy: partial cloud cover \\
    & W2 & Overcast: full cloud cover \\
    & W3 & Rain: rainfall conditions \\
    & W5 & Fog: reduced visibility due to fog \\
    & W8 & Night: after sunset until pre-dawn \\
    \hline
    \addlinespace
    \multirow{3}{*}{Risk Level} & High-risk & Actions likely to cause serious harm or property damage (E05, E11, E14, E17, E18, E19) \\
    & Moderate-risk & Actions that may result in moderate safety concerns or disruption (E06, E07, E09, E10, E12, E16) \\
    & Low-risk & Actions considered minor with minimal safety risk (E00, E01, E02, E03, E04, E08, E13, E15, E20) \\
    \bottomrule
  \end{tabular}
  \end{adjustbox}
\end{table}


\subsection{Annotation Workflow and Quality Assurance}

\textbf{Dataset Annotation. }
The A2Seek dataset employs a rigorous multi-level annotation framework to ensure high-quality and comprehensive labeling. As illustrated in Figure~\ref{fig:label_pipeline}, the annotation process begins with manually labeling anomaly categories and identifying relevant regions in keyframes, which are then extended to adjacent video frames using Grounded-SAM-2 \cite{ravi2024sam2segmentimages, ren2024grounding}. These annotations undergo cross-verification by human annotators to ensure consistency and accuracy. Based on this initial information, prompts were designed to guide a vision-language model in generating detailed annotations, including region-level, frame-level, and segment-level information. The model-generated annotations were further reviewed and refined by professional annotators to ensure alignment with human-labeled categories. \par
\begin{figure*}[t]
    \centering
    \includegraphics[width=\linewidth]{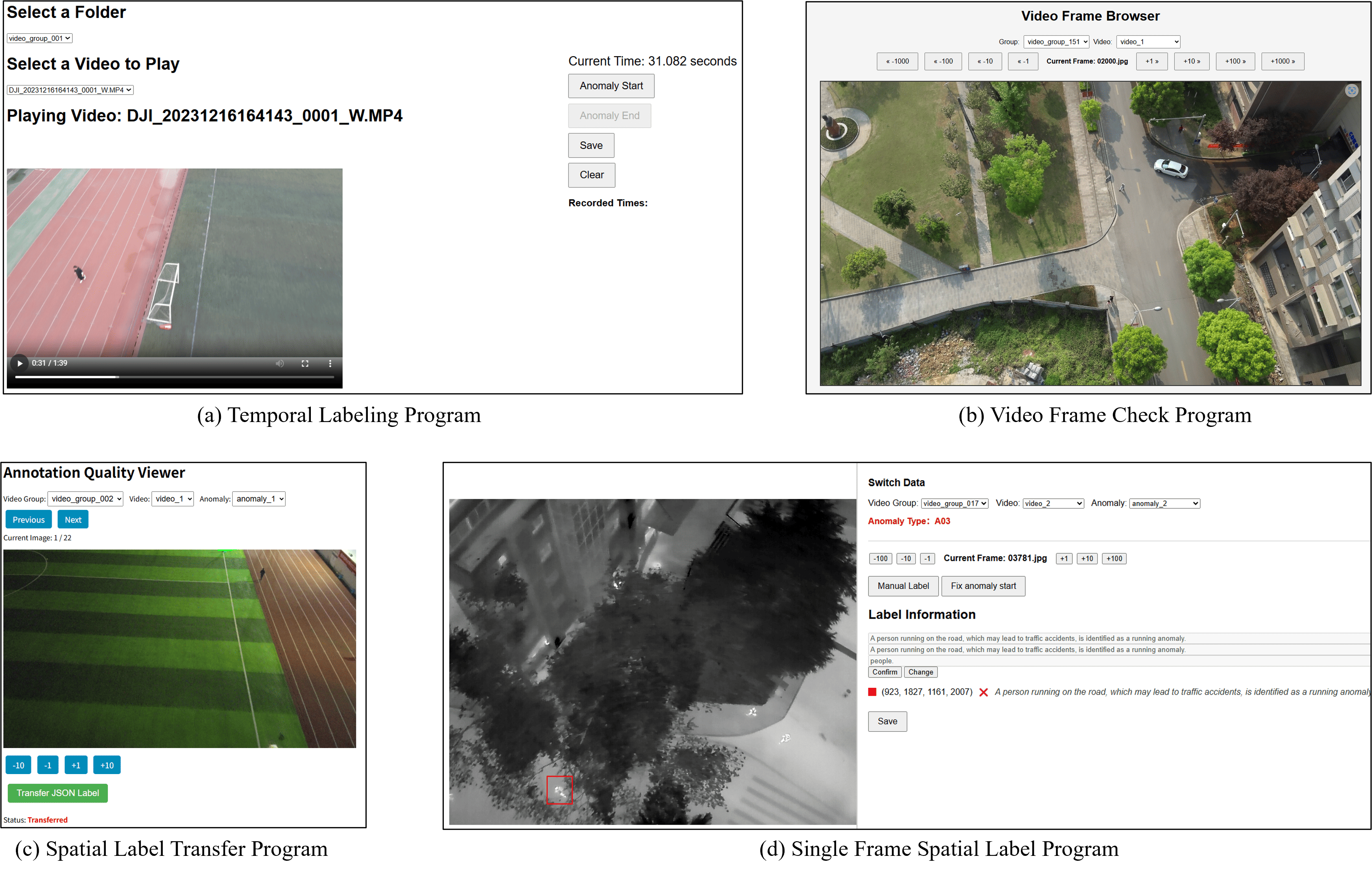}
    \caption{Screenshots of our in-house annotation toolkit.  
    \textbf{(a)} Temporal labeling interface: annotators record only the start frame, end frame and anomaly category for each event.  
    \textbf{(b)} Frame browser: allows rapid navigation to any frame in a clip so that timestamp alignment between raw videos and labels inside the same \textit{video group} can be visually verified.  
    \textbf{(c)} Label-transfer panel: human validators inspect SAM-generated masks and copy only approved instances into the final ground-truth store, filtering out noisy or redundant versions.  
    \textbf{(d)} Single-frame spatial labelling: experts draw a bounding box for every anomaly, attach a free-text description and supply key-words that drive the subsequent Grounding-SAM tracking step.}
    \label{fig:annotation_gui}
\end{figure*}

To enhance the reliability of the annotations, a multi-stage process was designed, encompassing trigger, diagnose, reasoning, reflection, and seeking phases. Specific formatting rules were employed to align the model outputs with human annotations, ensuring consistency and interpretability. During this process, the model was guided to generate multiple responses, from which professional annotators selected the most appropriate ones for further refinement. The finalized annotations were integrated into a reasoning-centric framework, providing structured explanations that include spatial localization, fine-grained category labels, and causal reasoning paths. This process ensures that the dataset not only supports precise anomaly detection but also facilitates in-depth semantic reasoning.
The annotation framework includes precise timestamps for each anomalous event, identifying the specific frame sequences where anomalies occur. Spatial localization is achieved through bounding boxes that accurately mark the positions of anomalous objects, ensuring high precision and reliability in anomaly detection. Additionally, natural language explanations describe the causes and contextual background of anomalies, forming dynamic reasoning paths based on visual scene content.
To maintain data quality, low-quality videos were removed, and privacy-sensitive information, such as faces and license plates, was processed to ensure compliance with ethical standards. This meticulous annotation process ensures that the A2Seek dataset provides a robust foundation for developing and evaluating aerial anomaly detection models in complex scenarios. 
Fifteen domain specialists spent roughly one month creating the A2Seek labels. The pipeline concentrates on three elements—\textit{temporal boundaries, spatial trajectories, and textual descriptions}—implemented with lightweight in-house tools (as shown in Figure~\ref{fig:annotation_gui}) and a round-robin verification scheme.  All outputs are released as COCO-VID-style JSON files.

\textbf{Temporal Boundaries. } 
We built a browser-based tool that lets an annotator scrub through each clip and press \textit{Start/End} buttons while selecting one of the 20 anomaly classes.  Every video is labelled independently by two experts; a third reviewer resolves mismatches.  The final time-stamps are exported to \texttt{temporal\_labels.json} together with the flight metadata recorded on location.

\textbf{Spatial Trajectories. }
For the first frame of each temporally marked event, an expert draws a bounding box and writes a short noun phrase (\textit{e.g.}, ``Bicycle'').  Two complementary tracking modes are then launched:

\begin{itemize}
\item BBox-only: the seed box is propagated frame-by-frame with a pretrained tracker.  
\item Keyword-guided: Grounding~DINO (Grounding DEtection with Improved deNoising anchor boxes)\cite{zhang2022dino} detects all boxes matching the phrase; the one with highest IoU to the seed is chosen, and Grounding-SAM (Grounding Segment Anything Model)\cite{ren2024grounded} refines and tracks the mask.  
\end{itemize}

The two results are merged, and a second annotator scrolls through the track to accept or flag it.  Flagged cases are re-seeded or re-edited until they pass inspection.   
Approved trajectories are stored in \texttt{spatial\_labels.json}.

\textbf{Textual Descriptions. }
Grounding-SAM's \texttt{\textless Caption\textgreater} head produces frame-level and clip-level captions. In addition, annotators provide a multi-steps description for every anomaly to capture intent and context. Captions are lightly proof-read before being written to \texttt{text\_labels.json}.

Unlike CoT-RAG \cite{li2025cot}, which models factual relationships between knowledge units, GoT focuses on structuring the \textit{reasoning trajectory}. This design allows adaptive control of reasoning depth in aerial-view anomaly understanding, where scene complexity and anomaly subtlety vary widely.

Graph-of-Thought (GoT) explicitly structures reasoning through \textbf{stage-specific tags} inserted into the generated sequence: 
\texttt{<|Trigger|>, <|Diagnose|>, <|Reasoning|>, <|Reflection|>, <|Seeking|>}.

Each stage represents a distinct cognitive operation:
\begin{itemize}
\item \textbf{Trigger}: Scene perception and anomaly signal activation.
\item \textbf{Diagnose}: Hypothesis formation for possible anomaly categories.
\item \textbf{Reasoning}: Contextual analysis and evidence-driven judgment.
\item \textbf{Reflection}: Self-evaluation or consistency check of reasoning.
\item \textbf{Seeking}: Optional localization or reference search for key regions.
\end{itemize}
Each stage is \textit{skippable}, allowing GoT to shorten reasoning chains for simple scenes and extend them for complex ones. We provide below two representative GoT annotations illustrating both simple and complex reasoning cases.

\begin{tcolorbox}[
    colback=gray!5,
    colframe=gray!80!black,
    title=GoT Annotation Example (a): Simple Scene (Normal),
    rounded corners,
    arc=2mm
]
\scriptsize
\ttfamily
\noindent\{\\
\ \ \ "trigger": "No individuals or objects are present in the scene. The environment consists of a running track, a soccer field, and an adjacent area with no visible activity or anomalies.",\\
\ \ \ "diagnose": ["Normal"],\\
\ \ \ "reasoning": "<no need>",\\
\ \ \ "reflection": "<no need>",\\
\ \ \ "seeking": "<no need>"\\
\}
\end{tcolorbox}

\begin{tcolorbox}[
    colback=gray!5,
    colframe=gray!80!black,
    title=GoT Annotation Example (b): Complex Scene (Lost Item),
    rounded corners,
    arc=2mm
]
\scriptsize
\ttfamily
\noindent\{\\
\ \ \ "trigger": "In the sequence of images, there is a noticeable change in the lower left area of the frame. Specifically, a person appears to be walking along a pathway near the edge of the lawn. This individual's presence was not evident in the initial frames but becomes apparent as the sequence progresses.",\\
\ \ \ "diagnose": "Possible anomaly categories include `Trespassing on Lawn' or `Lost Item'. The appearance of the person near the edge of the lawn could indicate an unusual presence, especially if they are not supposed to be in that area. Alternatively, the person might have dropped something, which would classify as a `Lost Item'.",\\
\ \ \ "reasoning": "The person is seen moving along a designated pathway, which suggests normal pedestrian behavior. However, their proximity to the lawn edge raises suspicion. Upon closer inspection, there is no direct indication of the person interacting with the lawn itself, such as stepping onto it. Instead, the focus shifts to the possibility of a lost item, as the person's movement does not suggest any abnormal interaction with the environment beyond walking.",\\
\ \ \ "reflection": "Given the lack of direct interaction with the lawn and the person's adherence to the pathway, the most plausible explanation is that the person has dropped an item. This aligns with the category 'Lost Item', as the person's movement pattern does not suggest trespassing but rather a search for something they may have misplaced.",\\
\ \ \ "seeking": "[270, 170, 582, 447]"\\
\}
\end{tcolorbox}

These examples illustrate how GoT represents reasoning as a \textit{semantic graph of thought stages}, bridging perception and interpretation in an interpretable and controllable manner.

\subsection{Dataset Features}
\paragraph{Anomaly Duration}
The anomalous events in the A2Seek dataset can be divided into three categories according to their duration: long-term, short-term, and instantaneous anomalies. 
Long-term anomalies are generally difficult to identify from a single frame and require extended contextual information for accurate judgment; such anomalies are often somewhat predictable, for example, prolonged loitering in restricted areas. 
Short-term anomalies last for a shorter period and typically require certain prior knowledge for detection, though some can still be recognized from individual frames; for instance, running across a pedestrian walkway may constitute a short-term anomaly. 
Instantaneous anomalies occur extremely rapidly, usually within only a few frames, making them difficult to detect. 
For instance, in theft scenarios, the entire action may occur in an instant, thereby complicating the identification process.

\begin{figure}[ht]
\centering

\begin{adjustbox}{max width=0.9\linewidth}
    \begin{tabular}{c|c c c c c c c c c c|c}
        \toprule
          & S00 & S01 & S02 & S03 & S04 & S05 & S06 & S07 & S08 & S09 & Sum\\
        \hline
        W0 & 6   & 31  & 23  & 22  & 52  & 32  & 5   & 13  & 17  & 0   & 201\\
        W2 & 12  & 9   & 5   & 35  & 37  & 25  & 2   & 5   & 2   & 0  & 132\\
        W3 & 0   & 18  & 1   & 0   & 8   & 29  & 5   & 8   & 0   & 0  & 69\\
        W1 & 2   & 3   & 3   & 18  & 14  & 1   & 0   & 5   & 0   & 1  & 47\\
        W5 & 0   & 0   & 0   & 0   & 0   & 0   & 0   & 0   & 0   & 6  & 6\\
        \hline
        Sum & 20 & 61 & 32 & 75 & 111 & 87 & 12 & 31 & 19 & 7 & 455\\
        \bottomrule
    \end{tabular}
\end{adjustbox}

\vspace{5pt}

\includegraphics[width=0.7\linewidth]{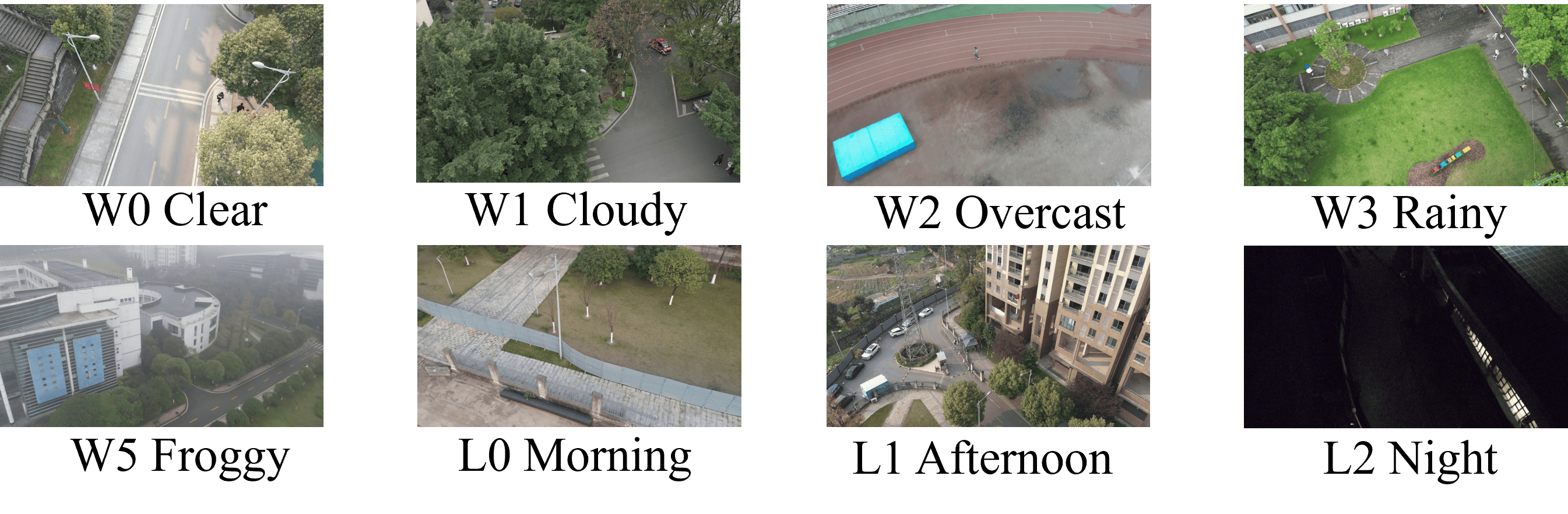}

\caption{Summary of weather, time, and scene types in our dataset. W0, W1, W2, W3, and W5 represent clear, cloudy, overcast, rainy, and foggy conditions, respectively, while S00 to S09 denote various scenes such as pathways and courtyards. The dataset covers recordings from morning, noon, and afternoon, excluding nighttime due to weather capture limitations. Out of 542 total videos, only 455 are included in the analysis.}
\label{fig:weather time scene}

\end{figure}

\begin{figure}[t]
  \centering
   \includegraphics[width=\linewidth]{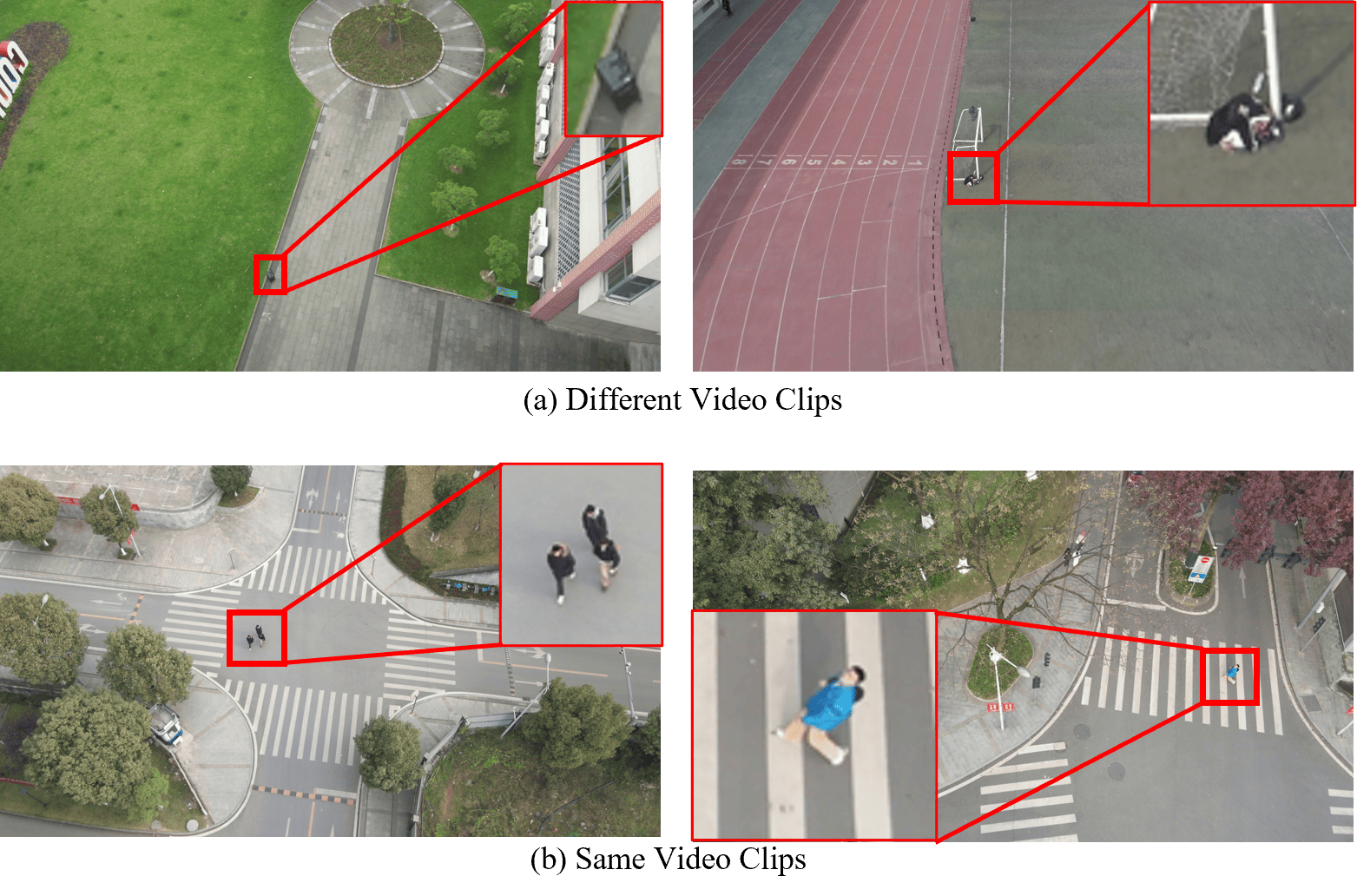}
   \caption{(a) Scene-dependent anomalies in different environments: An object left on the road is anomalous, while on the playground it is normal. (b) Scene-dependent anomalies in the same environment: Crossing outside a crosswalk is anomalous, while within it is normal.}
    \vspace{-2mm}
   \label{fig:scene dependence}
\end{figure}

\begin{figure}[t]
  \centering
   \includegraphics[width=\linewidth]{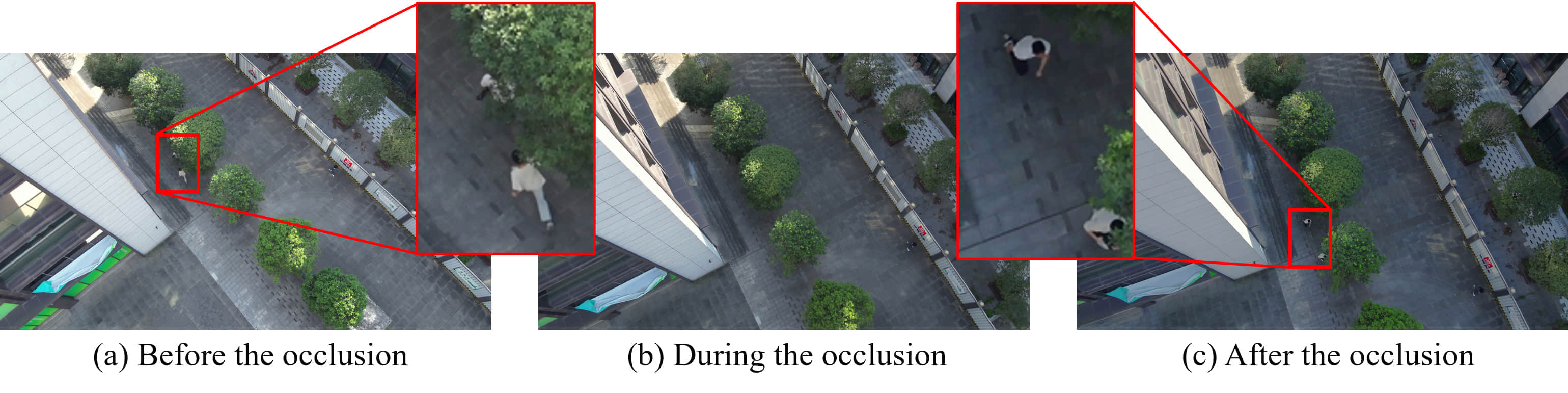}
   \caption{The left image shows two individuals chasing each other before entering the shadow of a tree. The middle image depicts the moment of occlusion, where they are fully obscured. The right image shows them emerging from the shadow after occlusion.}
    \vspace{-2mm}
   \label{fig:occlusion anomaly}
\end{figure}

\begin{figure}[t]
  \centering
   \includegraphics[width=\linewidth]{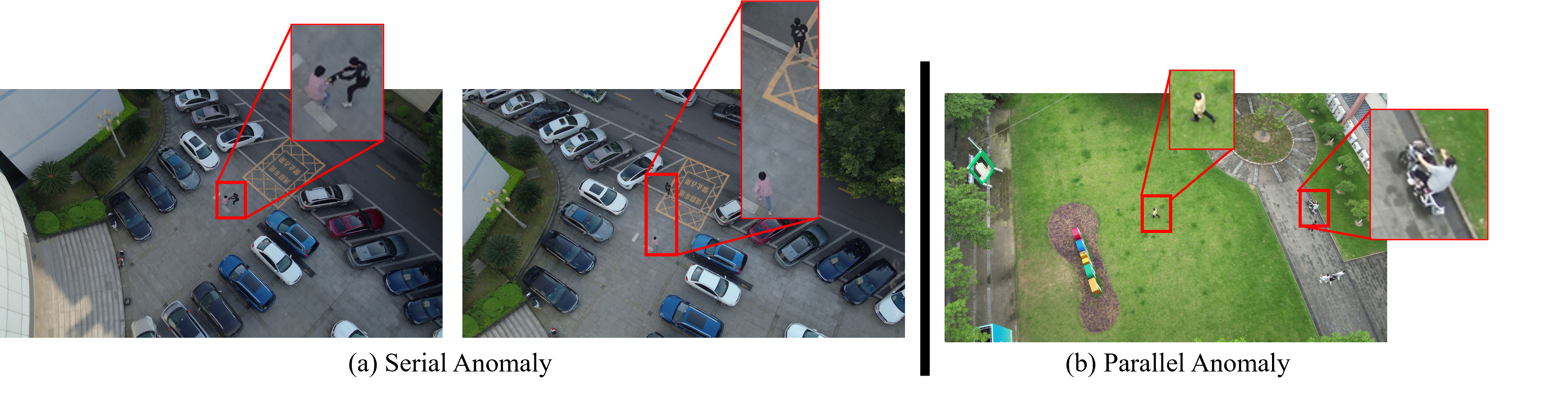}
   \caption{(a) shows a robbery followed by a chase, representing a serial anomaly. (b) depicts simultaneous anomalies at two locations, representing parallel anomalies.}
    \vspace{-2mm}
   \label{fig:serial and parallel anomaly}
\end{figure}

\paragraph{Scene-dependent Anomaly}

Scene-dependent anomalies are a significant feature of the A2Seek dataset, encompassing behavior anomalies that depend on specific scene contexts and occlusion anomalies caused by environmental obstructions. Scene-dependent anomalies can be categorized into two types: cross-video-segment scene-dependent anomalies, where a behavior that appears normal in one segment may be anomalous in another (as shown in Figure~\ref{fig:scene dependence}), and drone-specific scene-dependent anomalies, influenced by dynamic viewpoint shifts within the same segment (as shown in Figure~\ref{fig:occlusion anomaly}). Occlusion anomalies also frequently occur in drone views due to obstructions such as trees or buildings, posing unique challenges that require models to infer anomalies by leveraging temporal cues from preceding and succeeding frames. These complexities make drone-view anomaly detection substantially different from traditional surveillance settings, highlighting the importance of context-aware analysis. 


\paragraph{Serial and Parallel Anomalies}

Our dataset encompasses both serial anomalies and parallel anomalies. Serial anomalies refer to multiple anomalies occurring consecutively within the same time period, often exhibiting causal relationships, \textit{i.e.}, the occurrence of one anomaly triggers another. In contrast, parallel anomalies involve multiple anomalies occurring simultaneously within the same time period. These anomalies are generally independent of one another, lacking any causal connections. Figure \ref{fig:serial and parallel anomaly} demonstrate the serial and parallel anomaly in A2Seek dataset.




\section{Reward Design for Aerial Anomaly Understanding}

\label{appendix:reward_design}

\textbf{Format Reward}
To encourage structured and interpretable outputs, we introduce a format adherence reward. The model is required to organize its reasoning and final answer within optional intermediate steps, \textit{i.e.}, \texttt{<think>},  \texttt{<seeking>}, and \texttt{<answer>} tags. A reward of 1 is given for strict adherence to the template, while a reward of 0 is assigned if the format deviates. 

\textbf{Accuracy Reward.}
Let \(\hat{c}\) and \(c^{*}\) denote the predicted and reference anomaly classes extracted from the \texttt{<answer>} tag.  
The scalar reward \(R_{\text{cls}}\) is defined as  
\(R_{\text{cls}} = 1\) when \(\hat{c}=c^{*}\) (exact match);  
\(R_{\text{cls}} = 0.1\) when \(\hat{c}\neq c^{*}\) yet both classes are abnormal (\textit{i.e.}, \(\hat{c}\neq\textit{Normal}\) and \(c^{*}\neq\textit{Normal}\));  
and \(R_{\text{cls}} = 0\) in all other cases, including missing predictions.  
This scheme awards full credit for correct categorisation, partial credit for correctly flagging abnormality while misidentifying the subtype, and no credit for normal–anomaly confusion or omitted labels.

\textbf{Localization Reward.}
The localization reward \(R_{\text{loc}}\) is defined as the Intersection over Union (IoU) between the predicted bounding boxes \(\hat{B}\) and the ground truth boxes \(B^{*}\). 
This reward encourages the model to focus on the most relevant regions of the input, thereby improving its ability to identify and understand anomalies in complex aerial scenes.

\textbf{Seeking Reward.}
The seeking reward \(R_{\text{seek}}\) is designed to incentivize the model to determine the necessity of additional high-resolution context and, when required, to precisely localize the corresponding regions. Formally, it is expressed as 
\(R_{\text{seek}} = \mathbb{I}[\hat{s}=s^*] \cdot \text{IoU}(\hat{B}, B^*)\), 
where \(s^*\) and \(\hat{s}\) are the ground truth and predicted seeking decisions, and \(\text{IoU}(\hat{B}, B^*)\) measures the overlap between the predicted bounding box \(\hat{B}\) and the ground truth \(B^*\). 
This formula rewards the model for directly outputting predictions in simple scenarios, while encouraging it to simulate the process of focusing on suspicious areas by cropping and analyzing these regions for further inspection in complex scenarios.

\textbf{Length Reward.} To address the overthinking phenomenon \cite{sui2025stop} in simple scenarios, we couple answer correctness with reasoning length \(L\) (tokens in the \texttt{<think>} segments).  
If the answer is correct we favour brevity, setting {\small{\(R_d = \frac{1}{\log(1+L)}\)}}; if it is wrong we encourage elaboration with {\small{\(R_d = \min(L/L_{\max},\,1)\)}}, where \(L_{\max}\) equals the model's maximum output length.  
The reward is zero whenever either \texttt{<think>} or \texttt{<answer>} is missing.

\section{Discussed on the effectiveness of A2Seek-R1}
\label{appendix:discuss_effectiveness}

\subsection{Self-Correction via Reflection with Reasoning Annotations}

To further analyze how GoT-guided reasoning annotation in our A2Seek improves reasoning quality, we study model behavior with and without access to reasoning annotations.

\paragraph{Without GoT-guided Reasoning Annotations.} For input $x$, the model predicts $y_0$ based on a scoring function $s_\theta(x, y)$:
\begin{equation}
P_\theta(y \mid x) = \frac{\exp(s_\theta(x, y))}{\sum_{y'} \exp(s_\theta(x, y'))}.
\end{equation}
If $y_0$ is incorrect, there is no mechanism to revise it.

\paragraph{With GoT-guided Reasoning Annotations.} Reasoning annotations prompt the model to evaluate its own output and produce a refined prediction $\tilde{y}$:
\begin{equation}
P_\theta(\tilde{y} \mid x, y_0) \propto \exp\left(s_\theta(x, \tilde{y}) + \lambda \Delta r(x, y_0, \tilde{y})\right), 
\end{equation}
where $\Delta r(x, y_0, \tilde{y}) = R(x, \tilde{y}) - R(x, y_0)$ is the reflection-induced reward difference.

Assuming auxiliary rewards remain unchanged (\textit{e.g.}, localization), we approximate:
\begin{equation}
\Delta r(x, y_0, \tilde{y}) \approx R_{\text{cls}}(x, \tilde{y}) - R_{\text{cls}}(x, y_0).
\end{equation}
with
\begin{equation}
R_{\text{cls}}(x, y) = \mathbb{I}[y = y^*], \quad \Delta r(x, y_0, \tilde{y}) = \mathbb{I}[y_0 \neq y^* \wedge \tilde{y} = y^*] - \mathbb{I}[y_0 = y^* \wedge \tilde{y} \neq y^*].
\end{equation}

This reward difference yields a reflection-aware update:
\begin{equation}
\Delta \theta = -\eta \nabla_\theta \mathbb{E}[\Delta r(x, y_0, \tilde{y})].
\end{equation}

Finally, the improvement in expected classification reward is lower-bounded by:
\begin{equation}
\mathbb{E}[R_{\text{cls}}(\tilde{y})] - \mathbb{E}[R_{\text{cls}}(y_0)] \geq \eta \lambda \mathbb{E}[\Delta r].
\end{equation}

This analysis supports our empirical observation that reflection-guided self-correction leads to consistent gains in both accuracy and interpretability, and further highlights the value of structured annotations provided in A2Seek.
In summary, the reflection mechanism enables the model to revise suboptimal predictions by leveraging auxiliary hints and structured annotations, proving especially beneficial in scenarios with subtle or ambiguous anomalies. 
However, when the input information itself is insufficient (\textit{e.g.}, incomplete visual context or occlusions), self-correction alone may fail to imagine the missing clues. 
This motivates us to introduce a \emph{seeking} mechanism that actively queries for additional data, as detailed in the next section.

\paragraph{Seeking Mechanism.}
In scenarios where the input information is insufficient for accurate reasoning, the \texttt{<seeking>} mechanism allows the model to actively query for additional context, bridging the gap between the available input and the required information for correct predictions. To formalize this, we leverage the Information Bottleneck (IB) theory \cite{2024IBSurvey}, which balances the trade-off between the sufficiency of information for the task and the complexity of the representation.

Let \(I_{\text{input}}\) denote the information provided by the input \(x\), and \(I_{\text{required}}\) represent the total information needed for accurate reasoning. If \(I_{\text{input}} < I_{\text{required}}\), the model's reasoning process is under-constrained, leading to ambiguous or incorrect predictions. The \texttt{<seeking>} mechanism dynamically retrieves additional information \(I_{\text{seek}}\), such that the total information available becomes:
\[
I_{\text{total}} = I_{\text{input}} + I_{\text{seek}}.
\]

The seeking reward \(R_{\text{seek}}\) is designed to encourage the model to query for \(I_{\text{seek}}\) only when \(I_{\text{input}}\) is insufficient. Formally, the reward is defined as:
\[
R_{\text{seek}} = \begin{cases}
\beta \cdot \frac{I_{\text{seek}}}{I_{\text{required}}}, & \text{if } I_{\text{input}} < I_{\text{required}}, \\
0, & \text{otherwise},
\end{cases}
\]
where \(\beta\) is a scaling factor that controls the weight of the seeking reward. This formulation ensures that the model is incentivized to seek additional information only when it is necessary for accurate reasoning.

\paragraph{Information Bottleneck Perspective.}
From the perspective of the Information Bottleneck theory, the \texttt{<seeking>} mechanism can be viewed as a way to optimize the mutual information \(I(X; Y)\) between the input \(X\) and the output \(Y\), while minimizing the complexity of the intermediate representation \(Z\). The objective can be expressed as:
\[
\mathcal{L}_{\text{IB}} = I(X; Z) - \lambda I(Z; Y),
\]
where \(Z\) represents the information retrieved through \texttt{<seeking>}, and \(\lambda\) balances the trade-off between retaining sufficient information for the task and minimizing unnecessary complexity.

By incorporating the seeking reward \(R_{\text{seek}}\), the model dynamically adjusts \(I(Z; Y)\) based on the complexity of the input. For simple cases where \(I_{\text{input}} \approx I_{\text{required}}\), the model minimizes \(I(Z; Y)\) by avoiding unnecessary seeking. For complex scenarios where \(I_{\text{input}} \ll I_{\text{required}}\), the model increases \(I(Z; Y)\) by retrieving additional information, ensuring robust reasoning.

\paragraph{Unified Framework for Seeking and Reasoning.}
The seeking reward integrates seamlessly with the A-GRPO algorithm, influencing both the reflection reward \(\Delta r(x, y_0, \tilde{y})\) and the policy update. Specifically, the total reward \(R(x, y)\) now includes \(R_{\text{seek}}\) as a component:
\[
R(x, y) = R_{\text{format}} + R_{\text{acc}} + R_{\text{loc}} + R_{\text{length}} + R_{\text{seek}}.
\]

This unified framework ensures that the model balances exploration (querying for additional information) and exploitation (using the retrieved information to refine predictions). By dynamically adjusting the seeking process, the model achieves near-optimal performance across diverse and challenging UAV scenarios.

\section{Privacy Preservation, Licensing and Ethical Compliance}
\label{appendix:privacy}
\begin{figure}[htbp]
    \centering
    \includegraphics[width=\textwidth]{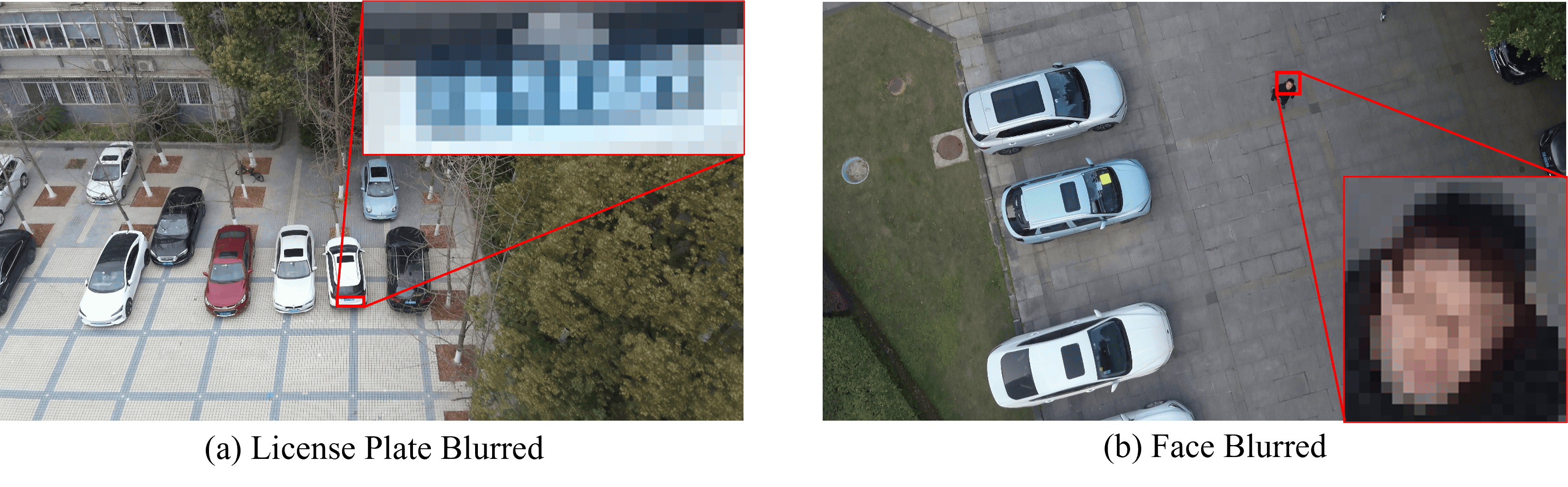}
    \caption{Examples of privacy protection in aerial video frames. 
    (a) License plates are pixelated to prevent vehicle identification. 
    (b) Facial regions are blurred to anonymize pedestrians. 
    These anonymization procedures ensure compliance with privacy-preserving standards during dataset release.}
    \label{fig:privacy_protection}
\end{figure}

Because human subjects are typically captured at oblique, metre-scale distances in the drone view, direct identification is already improbable; nevertheless, every frame is passed through an offline sanitisation pipeline. First, a lightweight YOLO-Face detector isolates facial regions, whose pixels are then scrambled with block-wise spectral noise in CIELab space and re-integrated via Poisson blending, leaving no reversible biometric cues while preserving motion and silhouette statistics for anomaly-detection research.

Although licence plates are rarely resolvable at the recorded altitudes, a human-in-the-loop audit team inspected every frame and manually applied the same irreversible scrambling procedure to any plate that remained decipherable. The final dataset and its metadata are distributed in read-only form, and access is gated by a click-through licence that explicitly forbids re-identification attempts and extends identical non-commercial, attribution requirements to all derivative models.

\section{Limitations, Open Challenges and Future Directions}

We introduce A2Seek—a large‐scale, multimodal UAV-view dataset enriched with dense bounding-box tracks and sentence-level descriptions—and A2Seek-R1, a multi-stage RL-fine-tuning framework that not only detects anomalies but also explains where and why they occur. To date, we measure performance with frame-level average precision and mean IoU, but our richly annotated data enable much more: future work will add region-aware metrics (\textit{e.g.}, temporal IoU of predicted vs. ground-truth tubes) and spatio-temporal localization scores that jointly reward correct timing and placement of anomaly predictions. Moreover, the moving UAV perspective—where objects shrink with altitude, perspective shifts continuously, and the camera itself pans—calls for bespoke evaluation criteria that penalize false alarms on distant clutter more heavily and incentivize early anticipation of emerging anomalies.

Methodologically, a UAV frame offers far richer cues than a static CCTV image. Parallax reveals 3D layout, rotor vibration modulates motion patterns, and onboard audio can flag sudden events like collisions or alarms. Building on this, forthcoming models will fuse high-resolution appearance with optical flow, depth priors and language grounding, while dynamically adapting their receptive fields as the drone zooms or sweeps its view. By mining these latent, multimodal signals, we aim to push anomaly detection beyond asking whether something is wrong toward explaining what is wrong, where it happens and why it matters.

\section*{Societal Impact}

Our proposed dataset focuses on abnormal event detection from UAV (Unmanned Aerial Vehicle) perspectives, with the primary goal of enhancing public safety and promoting robust perception models for long-range, real-time surveillance in open environments. This effort holds several \textbf{positive societal impacts}:

\begin{itemize}
    \item \textbf{Safety and early warning:} UAV-based monitoring allows timely detection of anomalous behaviors in large-scale outdoor areas (\textit{e.g.}, campuses, industrial zones), potentially preventing incidents such as violence or theft.
    \item \textbf{Technological advancement:} The dataset promotes the development of embodied multimodal AI models capable of spatial-temporal reasoning and active scene understanding.
    \item \textbf{Academic contribution:} It fills a gap in existing datasets by providing controlled abnormal scenarios with diverse contextual challenges, enabling reproducible and fine-grained benchmarking for the research community.
\end{itemize}

Despite these benefits, we recognize potential \textbf{negative societal impacts}, such as:

\begin{itemize}
    \item \textbf{Privacy concerns:} UAVs capture aerial footage, which may inadvertently include individuals or locations without consent.
    \item \textbf{Risk of misuse:} The dataset or associated models could potentially be adapted for surveillance beyond legitimate or ethical use cases.
\end{itemize}

To \textbf{mitigate these risks}, we take several strict measures:

\begin{itemize}
    \item \textbf{Ethical review and anonymization:} All collected data undergoes internal ethical review. No original raw video data is released; only extracted visual features and annotations are provided.
    \item \textbf{Identity protection:} All identifying visual information (\textit{e.g.}, faces, license plates, \textit{etc.}) is blurred or removed. No personal metadata is stored or shared.
    \item \textbf{Controlled data collection:} All actors performing abnormal events are volunteers from our research lab, who participated with full informed consent and received fair compensation.
    \item \textbf{Transparency and access control:} We provide access to visual features and metadata under the CC BY-NC-SA 4.0 license, strictly for academic, non-commercial research. Video URLs are included only to promote reproducibility, and content access remains subject to platform-level permissions.
\end{itemize}

We believe these efforts ensure that our work advances the field of multimodal abnormal event understanding in a safe, ethical, and socially beneficial manner.

\end{document}